\documentclass[journal]{IEEEtran}
\usepackage{enumitem}
\usepackage{multirow}
\usepackage{textcomp}
\usepackage{color}
\usepackage{booktabs}
\usepackage{threeparttable}
\usepackage{stfloats}

\usepackage{cite}
\ifCLASSINFOpdf
  \usepackage[pdftex]{graphicx}
\else
  \usepackage[dvips]{graphicx}
\fi

\usepackage{amsmath, amsthm, amssymb, amsfonts}
\usepackage{mathrsfs}
\usepackage[ruled, linesnumbered]{algorithm2e}  
\usepackage{algpseudocode}
\usepackage{array}
\usepackage{diagbox}

\ifCLASSOPTIONcompsoc
  \usepackage[caption=false,font=normalsize,labelfont=sf,textfont=sf]{subfig}
\else
  \usepackage[caption=false,font=footnotesize]{subfig}
\fi

\begin{document}
%
\title{A Hierarchical Deep Reinforcement Learning Framework for 6-DOF UCAV Air-to-Air Combat}
%
%
%

\author{Jiajun~Chai, ~\IEEEmembership{Student Member,~IEEE,}
        Wenzhang Chen, ~\IEEEmembership{Student Member,~IEEE,}
        Yuanheng~Zhu,~\IEEEmembership{Senior Member,~IEEE,}
        Zong-xin~Yao,
        and~ Dongbin~Zhao,~\IEEEmembership{Fellow,~IEEE,}

\thanks{This work was supported in part by the National Key Research and Development Program of China under Grant 2018AAA0102404, in part by the Strategic Priority Research Program of Chinese Academy of Sciences under Grant No. XDA27030400, and also in part by the National Natural Science foundation of China under Grant 62136008.}
\thanks{J. Chai, W. Chen, Y. Zhu, and D. Zhao are with the State Key Laboratory of Management and Control for Complex Systems, Institute of Automation, Chinese Academy of Sciences, Beijing 100190, China, and are also with the School of Artificial Intelligence, University of Chinese Academy of Sciences, Beijing 100049, China. Z. Yao is with the Shenyang Aircraft Design 
and Research Institute (SADRI). }
}

%
%

\markboth{Journal of \LaTeX\ Class Files,~Vol.~1, No.~1, January~2022}%
{Chai \MakeLowercase{\textit{et al.}}: A Hierarchical Deep Reinforcement Learning Framework for 6-DOF UCAV Air Combat}
%



\maketitle

\begin{abstract}
Unmanned combat air vehicle (UCAV) combat is a challenging scenario with continuous action space. In this paper, we propose a general hierarchical framework to resolve the within-vision-range (WVR) air-to-air combat problem under 6 dimensions of degree (6-DOF) dynamics. The core idea is to divide the whole decision process into two loops and use reinforcement learning (RL) to solve them separately. The outer loop takes into account the current combat situation and decides the expected macro behavior of the aircraft according to a combat strategy. Then the inner loop tracks the macro behavior with a flight controller by calculating the actual input signals for the aircraft. We design the Markov decision process for both the outer loop strategy and inner loop controller, and train them by proximal policy optimization (PPO) algorithm. For the inner loop controller, we design an effective reward function to accurately track various macro behavior. For the outer loop strategy, we further adopt a fictitious self-play mechanism to improve the combat performance by constantly combating against the historical strategies. Experiment results show that the inner loop controller can achieve better tracking performance than fine-tuned PID controller, and the outer loop strategy can perform complex maneuvers to get higher and higher winning rate, with the generation evolves. 
\end{abstract}


\IEEEpeerreviewmaketitle

\section{Introduction}
\IEEEPARstart{U}{nmanned} combat air vehicle (UCAV), an aircraft that can carry missiles for combat, is becoming more and more important in the future battlefield \cite{kumar2020brief}. Compared with manned aircraft, UCAV can greatly reduce the volume and weight of aircraft because it is not limited by the physical endurance of human pilots. The resulting lower cost and greater maneuverability make it an important member of the modern war system \cite{jordan2021future}. Air-to-air combat is a critical application scenario of UCAV, which can be categorized into Beyond Visual Range (BVR) combat and Within Visual Range (WVR) combat according to the combat range \cite{hu2022autonomous}. This study focuses on the WVR combat, in which the aircraft fights against the enemy at a close range. More specifically, an aircraft has a weapon engagement zone (WEZ) in front of it as the attack range of the aircraft guns or short-range missiles. The aircraft will be shot down if it stays in the WEZ of its enemy for a period of time. Each aircraft needs to attack enemies within its WEZ and avoid being attacked. It requires the aircraft to perform appropriate maneuver behaviors to gain more advantages. 

Intelligent air combat has attracted a lot of attention. The goal of intelligent WVR air combat is to control aircraft autonomously to shoot down enemy aircraft without being attacked. Since the UCAV controlled by human operators remotely may lose contact with the operators due to the electronic interference of the enemy \cite{li2021autonomous}, it is necessary to provide UCAV with intelligence for local decision-making. There have been studies trying to achieve intelligent air combat since 1950s \cite{isaacs1951games}. Early works focused on imitating the behavior of human pilots as expert experience. Some methods are based on several control rules \cite{arar2013flexible, chappell1992knowledge} or fuzzy logic \cite{ernest2016genetic, rao2011situation} to realize the experts experience. These methods face the problem that it is difficult to enumerate the behavior of experts in all cases, thus limiting their applications. 

In order to provide intelligence for aircraft without expert experience, a lot of works consider search-based or learning-based methods to obtain powerful combat strategies. For search-based methods, \textit{Park et al.} \cite{park2016differential} design a scoring function matrix to generate aircraft maneuver by min-max search. \textit{Mcgrew et al.} \cite{mcgrew2010air} adopt approximate dynamic programming to compute an efficient strategy. For learning-based methods, reinforcement learning (RL) is an appropriate solution to decision-making problems. It can learn the optimal decision strategies by interacting with environments instead of knowing the dynamics in advance \cite{shao2019survey}. Researchers have successfully applied this algorithm to some two-player zero-sum games \cite{zhu2020online}, such as Go \cite{silver2017mastering}, Atari games \cite{mnih2015human}, and StarCraft II \cite{vinyals2019grandmaster}. Besides, there are also some works that focus on multi-agent cooperation \cite{chai2021unmas, rashid2018qmix} and competition \cite{zhu2022empirical, zhang2021event, li2022missile}. Attracted by the performance of RL algorithms, there are a lot of works that adopt RL to train powerful strategies in WVR air combat. Many works adopt deep Q-learning \cite{mnih2015human} to solve the air combat task in discrete action space \cite{yang2019maneuver, xu2019autonomous, ma2018air}. \textit{Wang et al.} \cite{wang2020improving} adopt deep Q-learning to maintain the altitude and velocity of an aircraft. \textit{Yuan et al.} \cite{yuan2022research} propose a novel heuristic deep deterministic policy gradient (DDPG) algorithm to improve the exploration efficiency of the UCAV combat strategy in continuous action space. \textit{Weiren et al.} \cite{weiren2020air} combine multi-agent DDPG with game theory in one-on-one air combat. \textit{Li et al} \cite{li2022learning} adopt proximal policy optimization (PPO) to train the strategy fighting against a min-max strategy. Due to the model-free property of RL, there is no need for prior analysis or identification of the aircraft dynamics like traditional works \cite{seto2000case}. However, the methods above consider the air combat task under 3-DOF aircraft dynamics, which forces the velocity vector to be consistent with the nose direction. Therefore, the angle of attack and sideslip angle can be ignored. This simplification makes these methods unable to be applied to more realistic and complex 6-DOF aircraft dynamics. 

The movement of a real aircraft is governed by the 6-DOF dynamics, which consists of 12 nonlinear equations of motion \cite{heidlauf2018verification}. Direct training on 6-DOF dynamics becomes much more difficult than on 3-DOF dynamics, so researchers turn to hierarchical structures to reduce the training difficulty. The decision process is divided into two loops \cite{shin2018autonomous, wang2020influence, heidlauf2018verification}. The inner loop consists of the aircraft and the flight controller. It aims to implement the macro commands given by the outer loop, and manipulates the aircraft directly. The outer loop consists of the ono-on-one scenario and the combat strategy. It outputs the macro commands to the inner loop according to the relative observation of the other aircraft as well as own aircraft condition. \textit{Shin et al.} \cite{shin2018autonomous} adopt the nonlinear inverse control (NDI) \cite{snell1992nonlinear} to design a flight controller, and select predefined maneuvers according to a score function. \textit{Wang et al.} \cite{wang2020influence} also design an NDI-based flight controller to track the target flight-path angles, and use approximate dynamic programming to optimize intelligent combat strategies. However, the NDI-based control relies on the prior knowledge of aircraft dynamics, and in practice the control performance is vulnerable to internal and external disturbance. Many works also propose error-based \cite{zhou2015multi, porter1995genetic}, event-triggered \cite{li2020event}, and fault-tolerant \cite{abbaspour2018neural} flight controllers to avoid the dependence on aircraft dynamics. Most of them are studied on simple tasks, such as landing \cite{lin2015autolanding}, lateral control, or longitude control \cite{ashraf2018design}, and face obstacles in apply to air combat task where complex maneuvers are required. 

\textit{Pope et al.} \cite{pope2021hierarchical} design a different hierarchical framework and train the inner and outer loop only by RL. The inner loop consists of the aircraft and a sub-policy, which is trained by soft actor-critic (SAC) \cite{haarnoja2018soft}. They change the reward function to train several different sub-policies, which are intelligent enough to perform some basic maneuver behaviors with different preferences. Then, the outer loop represents a policy selector, which is also trained by SAC, to select an appropriate sub-policy for the current situation. This strategy eventually ranked the 2nd in AlphaDogfight Trials and defeated a human expert. However, the reward functions for sub-policies training are difficult to be well designed. Besides, the neural network used in the decision model is quite resource consuming, which requires ten of thousands of neurons on each layer.

\subsection{Contribution}
In this article, we focus on the WVR air-to-air combat under 6-DOF dynamics. Our goal is to present a powerful and model-free strategy under limited computing resources. We adopt the hierarchical framework to control the aircraft, and divide the decision process into the outer loop and inner loop, which maintain an air combat strategy and flight controller respectively. The main contributions are threefold: 
\begin{enumerate}
\item We propose a model-free hierarchical framework for 6-DOF air combat task. The combat strategy decides the target roll and pitch angles according to the current combat situation. Then, the flight controller tracks the target angles by calculating the actual input signals for the aircraft. Compared with other methods, the proposed framework requires less computing resources and its decision models do not require access to aircraft dynamics. 

\item We present an effective reward function by comprehensively balancing the control performance and aircraft constraints for the PPO algorithm with detailed analysis. Compared with traditional flight controllers, the decision process of our RL-based controller is model-free, and it can achieve better tracking performance in a wider range of scenarios. 

\item We improve the performance of the combat strategy by adopting the fictitious self-play mechanism to fight against the historical strategies generation by generation. In each generation, We find the approximate best response of historical strategies and store it into a pool. Experiment results show the good combating performance of our combat strategy.

\end{enumerate}

\subsection{Organization}
This article is organized as follows. Section II introduces the problem formulation of our air combat task and the PPO algorithm. Section III and section IV describe the design of inner loop flight controller and outer loop combat strategy in detail, respectively. Section V and section VI show the experiments and results of them. Finally, Section VII gives the conclusion.

\section{Problem Formulation}
\subsection{Air-combat Scenario}
\begin{figure}[htbp]
\centering
\includegraphics[scale=0.275]{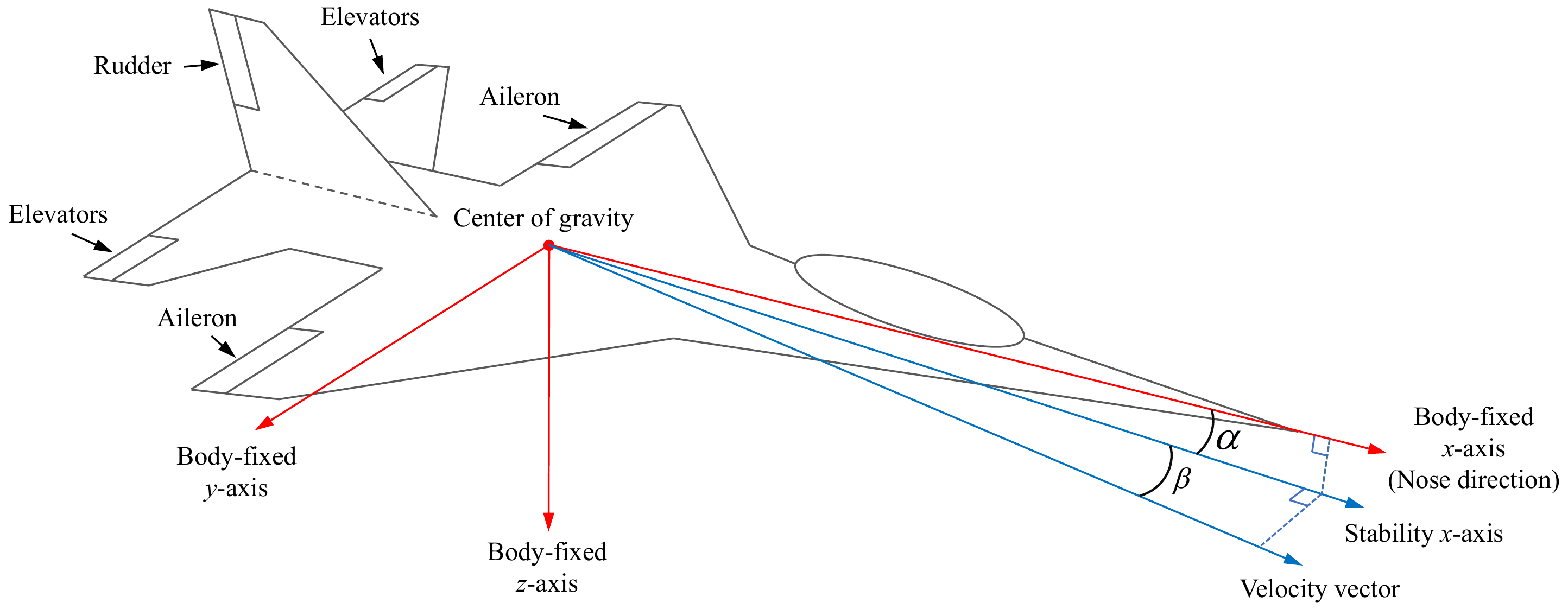}
\caption{Major components of 6-DOF dynamics. \cite{seto2000case}}
\label{fig:aircraft}
\end{figure}

In this paper, we choose F-16 as the combating aircraft, and its dynamics and aerodynamic parameters are all open-source. The first part of Table \ref{table:variable} shows the major state variables of 6-DOF aircraft. The motion of 6-DOF aircraft can be split into translational and rotational motions. For the translational motion, the aircraft changes its coordinates in the 3D space, which are northward displacement $pn$, eastward displacement $pe$, and altitude $alt$, respectively. The velocity vector shown in Fig. \ref{fig:aircraft} indicates the translational motion. The velocity of aircraft $vt$ is the 2-norm of the velocity vector: 
\begin{equation}
vt = \sqrt{\Dot{pn}^2 + \Dot{pe} ^ 2 + \Dot{alt} ^ 2}
\end{equation}
where $\Dot{pn}$, $\Dot{pe}$, and $\Dot{alt}$ are the derivatives of $pn$, $pe$, and $alt$ with respect to time, respectively. Besides, for the rotational motion, the Euler angles can represent the relationship between aircraft body-fixed coordinate system and ground coordinate system. $\phi$, $\theta$, $\psi$ are the Euler angle of roll, pitch, and yaw, respectively, and $P$, $Q$, $R$ are their derivative of time. 

\newcommand{\tabincell}[2]{\begin{tabular}{@{}#1@{}}#2\end{tabular}}
\begin{table}[htbp]
\centering
\caption{State and Action Variables of 6-DOF Aircraft \cite{heidlauf2018verification}}
\renewcommand{\arraystretch}{1.2}
\begin{tabular}{ccc}
\cline{1-3}
Variable 					& Bound	                  & Meaning 						       \\ \cline{1-3}
$v_t$        				& [0 m/s, 1.2 * 340 m/s]  & Velocity of aircraft				  \\
$pn$        				& -	                      & Northward displacement			  \\
$pe$        				& -	                      & Eastward displacement      		  \\
$alt$        				& [0 m, 12000 m]	      & Altitude of aircraft      		  \\
$\phi$, $\theta$, $\psi$ 	& [$-\pi$, $\pi$]	      & Angle of roll, pitch, and yaw		  \\
$P$, $Q$, $R$ 				& -                       & Rate of roll, pitch, and yaw		 \\
$\alpha$ 					& [$-\pi / 6$, $\pi / 3$] & Angle of attack					     \\
$\beta$ 					& [$-\pi / 6$, $\pi / 6$] & Angle of sideslip				         \\
$power$ 					& -	                      & Engine power lag				         \\
$\delta_{th}$ 					& [0, 1]                  & Actual engine throttle		     \\
$\delta_a$	                & [$-\pi / 9 $, $\pi / 9 $] & Actual aileron angle		         \\
$\delta_e$	                & [$-5 \pi / 36 $, $5 \pi / 36 $] & Actual elevator angle		         \\
$\delta_r$	                & [$-\pi / 6$, $\pi / 6$] & Actual rudder angle		         \\
\cline{1-3}
$\delta_{th}^{cd}$ 					& [0, 1]                  & Engine throttle	Command	     \\
$\delta_a^{cd}$	                & [$-\pi / 9 $, $\pi / 9 $] & Aileron angle	Command	         \\
$\delta_e^{cd}$	                & [$-5 \pi / 36 $, $5 \pi / 36 $] & Elevator angle Command		         \\
$\delta_r^{cd}$	                & [$-\pi / 6$, $\pi / 6$] & Rudder angle Command	         \\
\cline{1-3}
\end{tabular}
\label{table:variable}
\end{table}

Since the 6-DOF model allows the velocity vector to be different from aircraft nose direction, some more variables are added to the model. As shown in Fig. \ref{fig:aircraft}, the angle of attack $\alpha$ is the angle between nose direction and stability $x$-axis, which is the projection of velocity vector on the body-fixed X-Y plane. The angle of sideslip $\beta$ is the angle between stability $x$-axis and velocity vector. The motion of an aircraft is affected by the combined effects of gravity term, aerodynamics term, and engine power term. For the engine power term, the throttle of the engine $\delta_{th}$ determines the acceleration in the body-fixed $x$-axis. $power$ is the intermediate invariable of $\delta_{th}$ and represents the actual power of the engine. For the aerodynamics term, the aircraft has three control surfaces. $\delta_a$, $\delta_e$, $\delta_r$ are the deflection angle of aileron, elevator, and rudder, which are used to control the roll, pitch and yaw of the aircraft respectively. The second part of Table \ref{table:variable} shows the action variables of 6-DOF aircraft. The commands $u=\{\delta_{th}^{cd}, \delta_a^{cd}, \delta_e^{cd}, \delta_r^{cd}\}$ is different from the actual values shown in the first part. The state variables indicate the current deflection angles of the aircraft, while the variables with superscript $cd$ indicate the control commands to the aircraft. The state of an aircraft is governed by 12 nonlinear equations of motion \cite{heidlauf2018verification}. The dynamics can be simplified as:
\begin{equation}
\Dot{x} = f(x, u)
\end{equation}
where $x = [vt, alt, \alpha, \beta, \varphi, \theta, P, Q, R, pow, \delta_a, \delta_e, \delta_r]$ is the state of an aircraft, $u$ is the commands to the aircraft, $f(\cdot)$ is the nonlinear dynamic function for the aircraft motion. In this work, we formulate the control process of this continuous system as discrete control \cite{zhu2019invariant}. The details of the dynamics have been summarized by reference \cite{ahmed2019system, heidlauf2018verification}. 

\begin{figure}[htbp]
\centering
\includegraphics[scale=0.3]{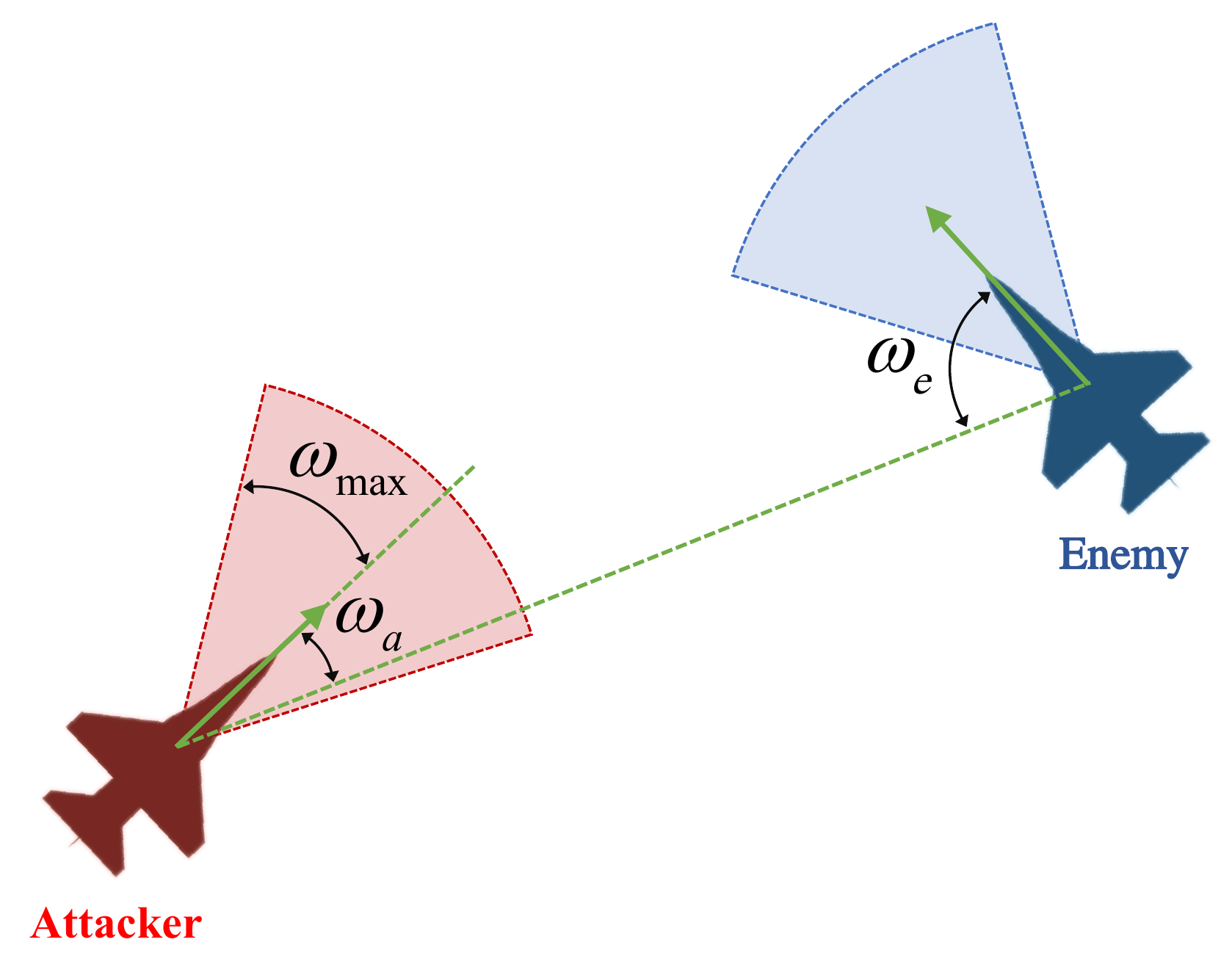}
\caption{Attack range of each aircraft in the combat scenario. }
\label{fig:attack-range}
\end{figure}

In this work, the battlefield is composed of two 6-DOF aircraft each of which carries infrared homing missiles to attack enemies. As shown in Fig. \ref{fig:attack-range}, the weapon engagement zone (WEZ) of the aircraft is a conical area. If an aircraft enters the WEZ of its enemy, it will be locked by the enemy missile and shot down. The cone angle $\omega_{max}$ of WEZ is 30 degrees and the generating line is 3 km. $\omega_{a}$ is the antenna train angle (ATA), which is the angle between line-of-sight (LOS) vector and the attacker aircraft nose direction. $\omega_e$ are the aspect angle (AA), which is the angle between LOS vector and the enemy aircraft nose direction. 

\subsection{Proximal Policy Optimization}
Reinforcement learning aims to get the optimal policy of the agent through the interaction with the environment. At each step of the interaction, the agent gets a state $s_k$ from the environment and determines an action $a_k$ for the state transition, where $k$ is the timestep. Then the environment will provide a reward $r_{k+1}$ to the agent to evaluate the action just being executed. The aim of RL is to maximize the sum of discounted rewards, which is also called return: 
\begin{equation}
\label{eq:return}
G_k = r_{k+1} + \gamma r_{k+2} + \gamma^2 r_{k+3} + ...
\end{equation}
where $\gamma \in [0, 1]$ is the discount factor that indicates history impact. 

Proximal policy optimization (PPO) is a widely-used RL algorithm based on policy gradient. It maintains two important modules: actor and critic, and both of them are generally represented by neural networks. The actor indicates the policy function $a \sim \pi(s;\Theta)$, which maps from states to action distribution probabilities. The critic $V(s;\Phi)$ approximates the state value function $V_\pi(s) = \mathbb{E}_\pi [\sum_{k=0}^K\gamma^k r_{k+1}|s_0=s]$, which is the expected return under policy $\pi$. $\Theta$ and $\Phi$ are parameters of two neural networks, respectively. The agent interacts with the environment and stores the state transition $\{s_k, a_k, r_k\}$ into a buffer. At the end of each episode, the returns $\{G_k\}_{k=0}^K$ of this episode are calculated according to Eq. (\ref{eq:return}) and stored into the buffer. The advantage value $A_k(s_k, a_k)$ is calculated by Generalized Advantage Estimator (GAE) and also stored into the buffer:
\begin{equation}
\begin{aligned}
&\hat{A}_k = \eta_k + (\gamma \lambda) \eta_{k+1} + ... + (\gamma \lambda)^{K-k-1}\eta_{T-1}\\
&where\ \ \eta_k = r_{k + 1} + \gamma V(s_{k+1};\Phi_{old}) - V(s_k;\Phi_{old})
\end{aligned}
\end{equation}
where $\hat{A}_k$ is the estimated value of $A_k(s_k, a_k)$, $\lambda$ is the hyper-parameter of GAE, $\Phi_{old}$ is the critic parameters when sampling these data, and $K$ is the timestep at the end of an episode. The advantage value is used to evaluate how much taking $a_k$ is better than the other actions. The PPO algorithm uses the stored data to optimize the critic and actor. The loss function of the critic is the mean squared error:
\begin{equation}
\label{eq:critic-loss-function}
L_{critic}(\Phi) = \mathbb{E}_k \left[V(s_t;\Phi) - G_t\right]^2.
\end{equation}

The learning objective of actor is maximizing:
\begin{equation}
\label{eq:actor-loss-function}
\begin{aligned}
L_{actor}(\Theta) = &\mathbb{E}_k \Big[ \min\Big(\rho_k(s_k, a_k;\Theta)\hat{A}_k,     \\
& clip[\rho_k(s_k, a_k;\Theta), 1-\varepsilon, 1+\varepsilon]\hat{A}_k\Big)\Big]
\end{aligned}
\end{equation}
where $\rho_k(s_k, a_k;\Theta)=\pi(a_k|s_k;\Theta) / \pi(a_k|s_k;\Theta_{old})$ is the probability ratio that measures the difference between $\pi(\Theta)$ and $\pi(\Theta_{old})$, where $\Theta_{old}$ is the actor parameters when sampling these data. This objective is modified from the surrogate objective $\mathbb{E}_k[\rho_k(a_k|s_k;\Theta)\hat{A}_k]$. Maximizing this surrogate objective may cause an excessively large policy update, making it difficult to accurately evaluate the updated policy on old data. Therefore, the objective uses the clip operation in Eq. (\ref{eq:actor-loss-function}) to avoid this by limiting the $\rho_k$ of new policy. $\varepsilon$ is the hyper-parameter to adjust the boundedness of the constraint. By minimizing Eq. (\ref{eq:critic-loss-function}) and maximizing Eq. (\ref{eq:actor-loss-function}), PPO can improve the policy performance and approach the optimal performance.

\section{RL-Based 6-DOF Flight Controller}
In the 6-DOF air combat task, a flight controller is necessary for the combat strategy to decrease its decision difficulty. With the help of flight controller, the combat strategy only needs to determine some macro behaviors, such as the aircraft attitudes, rather than directly determine the raw input signals for the aircraft. Some works make the flight controller to track assigned coordinates \cite{heidlauf2018verification, shin2018autonomous} to implement some basic maneuvers. However, the combat strategy based on these flight controllers cannot directly control the aircraft attitude, which limits the flexibility of maneuver. In this section, we design a flight controller that directly controls the aircraft attitude. It receives the target \emph{pitch} and \emph{roll} angle as tracking targets and outputs deflection angle commands of aileron, elevator, and rudder to control the aircraft directly. The Markov decision process (MDP) of the flight controller can be defined as a tuple $U = \{S, A, R, T, \gamma_f\}$, where $S$ is state space, $U$ is action space, $R$ is reward function, $T$ is state transition, and $\gamma_f$ is the discount factor. The details are as follows:

\begin{enumerate}[leftmargin = 1.5em]
\item[1)] \textbf{State}: The state $s_k$ for the flight controller at timestep $k$ describes the current situation of controlling the aircraft. It can be divided into two parts. The first part is the \emph{aircraft state} $x_k$, which has been introduced in section II.A. It contains enough variables to describe the aircraft current state for attitude control. The second part describes the current situation of tracking the target pitch and roll angle. It contains the tracking error of pitch and roll angle: $[\Tilde{\theta}_k - \theta_k, \Tilde{\varphi}_k - \varphi_k]$, where $\Tilde{\theta}_k$ and $\Tilde{\varphi}_k$ are the target pitch and roll angle, respectively. In conclusion, state $s_k$ is $[m_k, \Tilde{\theta}_k - \theta_k, \Tilde{\varphi}_k - \varphi_k]$. Besides, all bounded variables are normalized according to their upper and lower bounds. 

\item[2)] \textbf{Action}:  The action $a_t$ contains the deflection angle commands of aileron $\delta_a^{cd}$, elevator $\delta_e^{cd}$, and rudder $\delta_r^{cd}$, which can control the roll, pitch and yaw of the aircraft, respectively. 

\item[3)] \textbf{Reward}: The reward is one of the most important elements in the RL training process. For the training of flight controller, we design an effective reward function as follows:
\begin{equation}
r_k = c_1 r_k^{alive} + c_2 r_k^{track} + c_3 r_k^{action}.
\end{equation}

The first part $r_k^{alive}$ is an alive bonus. Its assignment rule is as follows:
\begin{equation}
r_k^{alive} = 
\begin{cases}
+1, & alive \\
0,  & instability \ or \ crash
\end{cases}
\end{equation}

If the aircraft does not fail due to instability or crash, it will receive a positive reward. The second part $r_k^{track}$ is the tracking reward, which is the negative mean square error between the target angles and the current pitch and roll angle:
\begin{equation}
r_k^{track} = - \sqrt{(\Tilde{\theta}_k - \theta_k)^2 + (\Tilde{\varphi}_k - \varphi_k)^2}
\end{equation}
where $\Tilde{\theta}_k$ and $\Tilde{\varphi}_k$ are the target pitch and roll angle, respectively. The third part $r_k^{action}$ is the action reward that is helpful to generate smooth control signals and reduce system fluctuation. It is the negative square sum of actions to encourage the controller to control the aircraft smoothly:
\begin{equation}
r_k^{action} = - \sqrt{(\delta_a^{cd})^2 + (\delta_e^{cd})^2 + (\delta_r^{cd})^2}
\end{equation}
We can adjust the coefficients $c_1$, $c_2$, and $c_3$ of $r_k$ to change the effect of these parts. We will discuss it in the experiment.

\item[4)] \textbf{State Transition}: The transition function $T$ represents the probability distribution ${\rm Pr}(s^\prime|s, a; \delta_{th}^{cd})$ of the next state $s^\prime$. For the aircraft, its state transition can be described by $x_{k+1} = \int_{k*T_f}^{(k+1)*T_f}{f\left(x(\tau), [a_k, \delta_{th}^{cd}]\right)}{\rm d \tau}$ , where $\delta_{th}^{cd}$ is the throttle command that is an external variable for the flight controller MDP. Besides, there is a decision interval $T_f$ between two adjacent decisions, because we formulate the control process as discrete control. 

\end{enumerate}

\begin{algorithm}[t]
  \caption{PPO-based Flight Controller Training Algorithm}
  \label{alg:controller}
  Initialize parameters of actor $\Theta_f$ and critic $\Phi_f$\; 
  \For {$epoch = 1 \ to \ m$}
  {
    Initialize the replay buffer $\mathcal{M}_{f}$\;
  	\While {$\mathcal{M}_{f}$ is not full}
  	{
  		Initialize the target signal randomly\;
  		\For {$t = 1 \ to \ T$}
  		{
  			Re-initialize target signal at a fixed period\;
  			Get micro state $s_k$\; 
  			Update the target signal\;
  			Take micro action $a_k$ by $\sigma(s_k;\Theta_f)$\;
  			Execute action and get micro reward $r_k$\; 
  			Store $\{s_t, a_k, r_k\}$ into $\mathcal{M}_f$\; 
		}
  		Compute $\hat{A}^f$ and $G^f$ and store them to $\mathcal{M}_f$\;
	}
  	\For {$Update \ times$}
  	{
  		Update $\Theta_f$ by maximizing Eq. (\ref{eq:actor-loss-function})\;
  		Update $\Phi_f$ by minimizing Eq. (\ref{eq:critic-loss-function})\;
	}
  }
\end{algorithm}

The control process with state transition is summarized in the inner loop part of Fig. \ref{fig:framework} (b). The flight controller maintains a policy $a_k \sim \sigma(\cdot|s_k;\Theta_f)$, which is represented by a neural network with $\Theta_f$ as its parameter. It can output action $a_k$ according to the state $s_k$. The flight controller receives the target pitch $\Tilde{\theta}$, and roll $\Tilde{\varphi}$ angle to calculate the tracking errors, and then combines it with the aircraft state $x$ to form micro state $s$. Then, the flight controller outputs deflection angle commands $a = [\delta_a^{cd}, \delta_e^{cd}, \delta_r^{cd}]$ according to $s$. Finally, the aircraft combines $a$ and the external command $\delta_{th}^{cd}$ and changes to the next state $s^\prime$. 


The flight controller is trained by PPO algorithm, and we present the training algorithm in Algorithm \ref{alg:controller}. The parameters of the actor and critic network for flight controller are initialized as $\Theta_f$ and $\Phi_f$. At the beginning of each epoch, we need to initialize a replay buffer $\mathcal{M}_f$ and start sampling. At the beginning of each episode, one aircraft is initialized at a legal state. Then, the controller tries to execute the macro commands. These commands come from two target signals, which are the sequence of roll and pitch angles. For a fixed length timesteps, the target signals will be re-randomized to create a diversity of training scenarios. Besides, the external command $\delta_{th}^{cd}$ is also initialized randomly in the training of flight controller, so that the controller can be more stable in all situations. The flight controller tries to track these signals and generates many state transitions $\{s_k, a_k, r_k\}$ to be stored into $\mathcal{M}_f$. At the end of each episode, we compute the estimated advantage values $\hat{A}^f$ and returns $G^f$ and store them into $\mathcal{M}_f$. When enough data is sampled, we update the actor parameters $\Theta_f$ and critic parameters $\Phi_f$ according to the loss functions described in section II.B. By repeating the training algorithm, the PPO-based flight controller can converge and achieve good tracking performance, which will be verified in experiments.

\section{Hierarchical Decision for 6-DOF Air Combat}
In this section, we design a combat strategy on the basis of the flight controller. We divide the control process into inner loop and outer loop. The flight controller is used as the inner loop controller. The combat strategy controls the outer loop. It outputs the target angle rates to generate target signals for the flight controller according to the combat situation that consists of the state of two aircraft. Besides, we also propose the training algorithm of the combat strategy and combine it with the RL-based flight controller. The MDP of combat strategy can be defined as a tuple $\mathscr{U} = \{\mathbb{S}, \mathbb{A}, \mathbb{R}, \mathbb{T}, \gamma_c\}$. To avoid confusion, we use $\mathfrak{s}_k$, $\mathfrak{a}_k$, and $\mathfrak{r}_k$ to distinguish from $s_k$, $a_k$, and $r_k$ of the flight controller. The details are as follows: 

\begin{algorithm}[ht]
  \caption{Training Algorithm of 6-DOF Combat Strategy via Self-Play}
  \label{alg:strategy}
  Initialize an historical strategy pool $\Pi_0$\; 
  Get a flight controller $\sigma(\cdot;\Theta_f)$ by Algorithm \ref{alg:controller}\; 
  \While {combat strategy is not convergence}
  {
    Initialize parameters of actor $\Theta_c$ and critic $\Phi_c$\; 
    \While {not converge}
    {
      Initialize the replay buffer $\mathcal{M}_{c}$\;
    	  \While {$\mathcal{M}_{c}$ is not full}
  	  {
  		Initialize two aircraft in the scenario\;
  		Sample an strategy from $\Pi_\ell$ in random\;
  		\While {episode does not end}
  		{
  			Get macro state $\mathfrak{s}_k$ from environment\; 
  			Take macro action $\mathfrak{a}_k$ by $\pi(\mathfrak{s}_k;\Theta_c)$\;
  			Feed $\mathfrak{a}_k$ to the flight controller\;
			\For {$j = 1 \ to \ T_c/T_f$}
			{
			  Get micro state $s_j$\; 
			  Update target signal by $\mathfrak{a}_k$ and $s_j$\;
			  Get micro action $a_j$ by $\sigma(s_j;\Theta_f)$\;
  		    Execute micro action $a_j$ and get current macro reward $r_j$\; 
  			}
  			Get macro reward $\mathfrak{r}_k = \sum_j r_j$\;
  			Store $\{\mathfrak{s}_k, \mathfrak{a}_k, \mathfrak{r}_k\}$ into $\mathcal{M}_c$\; 
		}
  		Compute $A^c$ and $G^c$ and store them to $\mathcal{M}_c$\;
	}
  	  \For {$Update \ times$}
  	  {
  		Update $\Theta_c$ by maximizing Eq. (\ref{eq:actor-loss-function})\;
  		Update $\Phi_c$ by minimizing Eq. (\ref{eq:critic-loss-function})\;
	  }
    }
    Store the converged strategy into the pool $\Pi_{\ell+1} = \Pi_\ell \cup \{\pi(\cdot;\Theta_c)\}$;
  }
\end{algorithm}

\begin{enumerate}[leftmargin = 1.5em]
\item[1)] \textbf{State}: The state $\mathfrak{s}_k$ comes from the combat scenario and consists of two parts. The first part is the coordinate of the enemy aircraft in the body-fixed coordinate system of the ego aircraft as follows:
\begin{equation}
p_{re} = M [pn - pn^\prime, pe - pe^\prime, alt - alt^\prime]
\end{equation}
where $pn^\prime$, $pe^\prime$, and $alt^\prime$ are the coordinate of the enemy aircraft, and $M$ is the transformation matrix from the earth coordinate system into the body-fixed coordinate system of ego aircraft:
\begin{equation}
M = 
\begin{bmatrix}
\cos\theta\cos\varphi   & \cos\theta\sin\varphi     & \sin\theta    \\
-\sin\varphi            & \cos\varphi               & 0             \\
\sin\theta\cos\varphi   & \cos\theta\sin\varphi     & \cos\theta    \\
\end{bmatrix}
.
\end{equation}

The second part contains the following state variables $[vt, alt, \alpha, \beta, \varphi, \theta, P, Q, R, pow]$ to describe the aircraft current condition. 

\item[2)] \textbf{Action}: The action $\mathfrak{a}_k$ has two parts. The first part is the throttle command $\delta_{th}^{cd}$ that is directly provided to the aircraft. The second part is the \emph{angel rates} of roll $\Dot{\varphi}$ and pitch $\Dot{\theta}$. They are used to calculate the target (roll $\Tilde{\varphi}$ and pitch $\Tilde{\theta}$) angles from the current angles to the bounds. 

\item[3)] \textbf{Reward}: The reward $\mathfrak{r}_k$ is sparse, which only occurs at the end of an episode. If the ego aircraft exceeds the altitude limitation or is shot down by the enemy, it will receive a negative penalty, whose value is -100. If the ego aircraft shoots down the enemy, it will win this combat and receives a positive reward, whose value is +100. 

\begin{figure*}[htbp]
\centering
\includegraphics[scale=0.475]{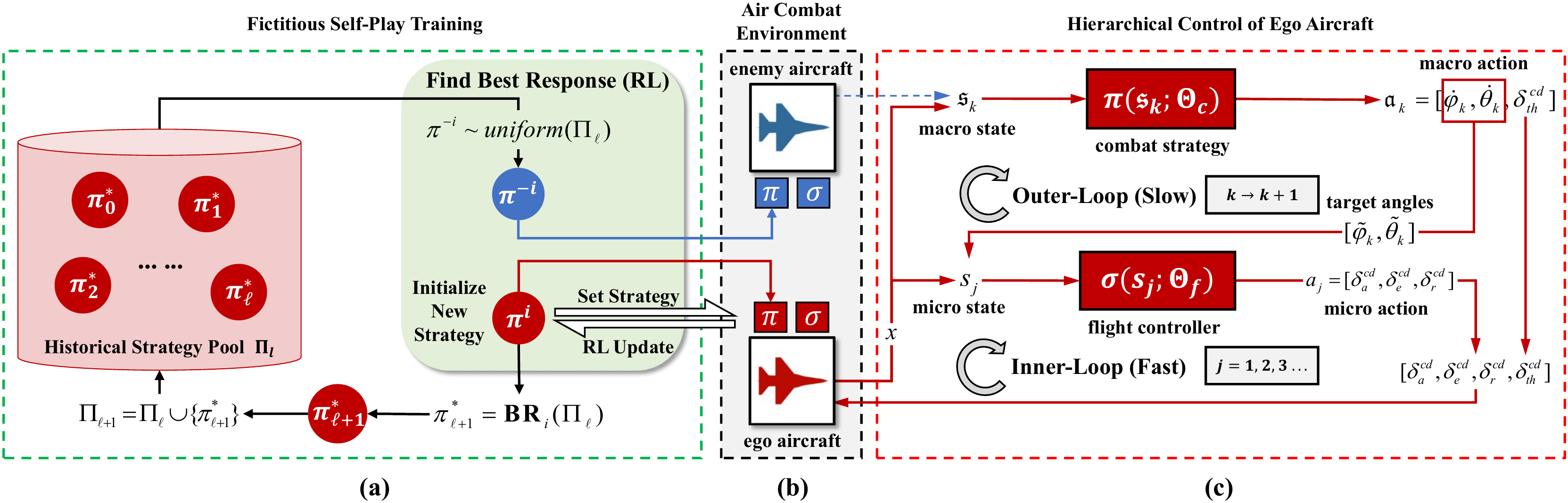}
\caption{Hierarchical framework for air combat. (a) The fictitious self-play mechanism to improve the performance of combat strategy. (b) The air combat environment that consists of two 6-DOF aircraft. (c) The hierarchical control framework for 6-DOF aircraft. We consider the red aircraft as the ego aircraft, and describe its control and training process. }
\label{fig:framework}
\end{figure*}

\item[4)] \textbf{State Transition}: The transition function $\mathbb{T}$ of the combat strategy can be defined as ${\rm Pr}(\mathfrak{s}^\prime|\mathfrak{s}, \mathfrak{a})$. The decision interval of the combat strategy is $T_c$. As shown in Fig. \ref{fig:framework} (c), at each decision of the outer loop, the combat strategy outputs action $\mathfrak{a}$ according to state $\mathfrak{s}$. According to action $\mathfrak{a}$, the inner loop calculates the target angles $\Tilde{\theta}$ and $\Tilde{\varphi}$ and combines them with throttle command $\delta_{th}^{cd}$ to form the macro command. The flight controller repeats the inner loop control for $T_c$ to perform the macro command. The enemy also moves to a new state. Then, an outer loop is completed, and the new state $\mathfrak{s}^\prime$ is reached. Since frequently changing the macro commands may cause the instability of flight controller, the target signals and throttle command will be reset after a larger decision interval $T_c$.

\end{enumerate}

The combat strategy maintains a policy $\mathfrak{a}_k \sim pi(\cdot|\mathfrak{s}_k;\Theta_c)$, which is also represented by a neural network with $\Theta_c$ as its parameter. In the training of combat strategy, we adopt the fictitious self-play mechanism to improve the combat performance from scratch. The one-on-one air combat task can be formalized as a zero-sum game, meaning that the reward summation of two players is always zero:
\begin{equation}
\mathfrak{r}_k^i + \mathfrak{r}_k^{-i} = 0, \forall t
\end{equation}
where $\mathfrak{r}_k^i$ and $\mathfrak{r}_k^{-i}$ indicate the reward of ego player and the opponent, respectively. The fictitious self-play mechanism makes the player optimize its current strategy by combating against its historical strategies. Therefore, we establish a historical strategy pool $\Pi$ to record the historical strategies, and store the \emph{level straight flight} strategy as the first strategy. This simple strategy keeps the initial altitude of the aircraft and flies at the same direction without other maneuvers. 


As shown in Fig. \ref{fig:framework} (a), in each generation, the player needs to find the best response against historical strategy pool $\Pi_\ell$:
\begin{equation}
\pi^*_{\ell+1} = \textbf{BR}_i (\Pi_\ell)
\label{eq:best-response}
\end{equation}
where $\Pi_\ell$ is the historical strategy pool of $\ell$-th generation, $\textbf{BR}_i (\Pi_\ell)$ is the best response against $\Pi_\ell$ for player $i$. For the 6-DOF WVR air combat, it is difficult to find the accurate best response against the historical strategy pool. We employ PPO algorithm to find the approximate best responses. We train the combat strategy until the PPO algorithm is converged. The training process is shown in Algorithm \ref{alg:strategy}. The training is similar with Algorithm \ref{alg:controller}. At the beginning of each episode, the strategy of player $-i$ is sampled uniformly from $\Pi_\ell$:
\begin{equation}
\pi^{-i} \sim uniform(\Pi_\ell).
\end{equation}
Then the combat strategy fights against its enemy. It should be noted that the combat strategy is used in the outer loop, so several inner loops need to be executed between two macro decisions as shown in lines 14 to 19. Besides, $\hat{A}^c$ and $G^c$ in lines 23 represents the estimated advantage values and returns for the combat strategy in the last episode. The best response $\textbf{BR}_i (\Pi_\ell)$ trained by PPO means that $\pi^*_{\ell+1}$ is the strategy that can get the largest reward against the uniformly sampled historical strategy pool. The converged best response strategy after a generation will be added to $\Pi_\ell$ to form a new historical strategy pool and start a new generation until convergence:
\begin{equation}
\Pi_{\ell+1} = \Pi_\ell \cup \{\pi^*_{\ell+1}\}. 
\end{equation}
The fictitious self-play process is considered to reach convergence when the new strategy shows little improvement over the historical strategies in $\Pi$.

\section{Experiments on RL-based Flight Controller}
\subsection{Experimental Setup}
In the experiments, we set the decision interval $T_f$ of the flight controller to 0.04 seconds. For the RL training experiment, we apply PPO algorithm to optimize the flight controller. The rule of generating the random target signals in Algorithm \ref{alg:controller} can be formulated as follows:

\begin{enumerate}[leftmargin = 1.5em]
\item[a)] \textbf{Randomize stable angle}: We first get a random value $\mathscr{Y}$ in the value range of pitch and roll as shown in Table \ref{table:variable}. The target signal will eventually stabilize at this value. 

\item[b)] \textbf{Randomize angle rate}: In order to avoid generating target
signals with sharp changes, we randomize the angle rate $\Dot{y}$ within a value range, instead of generating a step signal. If $\mathscr{Y}$ is greater than the initial angle $y_0$, the angle rate takes random value in the positive value range, otherwise in the negative value range. The value range is $[-0.2\ rad / s, 0.4\ rad / s]$ for pitch angle rate, and $[-0.8\ rad / s, 0.8\ rad / s]$ for roll angle rate. 

\item[c)] \textbf{Generate target signal}: The target signal $\Tilde{y}$ starts from the initial angle $y_0$ and follows the angle rate $\Dot{y}$ \textbf{until} the stable angle is reached:
\begin{equation}
\Tilde{y} = 
\begin{cases}
clip(y_0 + \Dot{y} * (t - t_0), -\infty, \mathscr{Y}) & \Dot{y} > 0 \\
y_0 & \Dot{y} = 0 \\
clip(y_0 + \Dot{y} * (t - t_0), \mathscr{Y}, +\infty) & \Dot{y} < 0
\end{cases}
\label{eq:random-signal}
\end{equation}
where $y$ can be the angles of pitch $\theta$ and roll $\phi$. $t$ and $t_0$ are the current and initial time, respectively. 

\end{enumerate}

Besides, since the three parts of flight controller reward have different effects on the performance of controller, we conduct several experiments to find the coefficients pair with best performance. We set $c_1$ to 1.0, and adjust $c_2$ and $c_3$. The coefficient $c_2$ of the tracking penalty varies from 0.5 to 2.5, and the coefficient $c_3$ of the action penalty varies from 0.0 to 0.2. In the experiments of evaluating the tracking performance of flight controller, we show the tracking curves of the roll and pitch angles under three target signals. The aircraft is initialized to level flight state, and the description of the three signals are as follows: 
\begin{enumerate}[leftmargin = 1.5em]
\item[a)] \textbf{Sine-cosine Signal}: We use the sine-cosine signal to test the performance of flight controller under smooth changes:
\begin{equation}
\Tilde{y} = \mathscr{A}\sin(\frac{2\pi}{\mathscr{T}}*t)
\end{equation}
where $\mathscr{A}$ and $\mathscr{T}$ is the amplitude and period of the signal, respectively. We set $\mathscr{A}$ to 15 degrees and $\mathscr{T}$ to 10 seconds. 

\item[b)] \textbf{Step Signal}: We use the step signal to test the performance of flight controller under sharp changes:
\begin{equation}
\Tilde{y} = 
\begin{cases}
\mathscr{A} & {\rm int}(t / \mathscr{T}) \ mod \ 2 = 0 \\
0 & {\rm int}(t / \mathscr{T}) \ mod \ 2 = 1 \\
\end{cases}
\end{equation}
where ${\rm int}(\cdot)$ is the rounding down operation, and $mod$ is the remainder operation. We set $\mathscr{A}$ to 15 degrees and $\mathscr{T}$ to 10 seconds. 

\item[c)] \textbf{Random Signal}: The random signal is in the same form as Eq. (\ref{eq:random-signal}). It is suit for the macro behavior of our RL-based combat strategy. We set the signal re-randomized every 5 seconds. 

\end{enumerate}

Besides, in order to compare the performance of our RL-based flight controller, we design a proportional integral derivative (PID) controller to track the target roll and pitch angle. It is a classical controller that outputs commands based on tracking error \cite{zhou2015multi}. Its control law is shown as follows:
\begin{equation}
\label{eq:pid}
a(t) = K_p e(t) + K_i \int_0^t e(\tau){\rm d}\tau + K_d\frac{{\rm d}e(t)}{{\rm d}t}
\end{equation}
where $a(t)$ is the commands, $e(t)$ is the tracking error. The three parts on the right side of Eq. (\ref{eq:pid}) are the proportional, integral and differential term of error. They represent the magnitude, accumulation, and trend of the tracking error, respectively. $K_p$, $K_i$, and $K_d$ are their coefficients, which are generally adjusted according to empirical experience to achieve satisfying results. In the 6-DOF aircraft dynamics, the deflection angle of aileron and elevator controls the angle of roll and pitch, respectively \cite{qiao2022morphing}. Besides, since the effect of the deflection angle of rudder is small, we ignore it without any PID controller and set the rudder to a fixed 0. Thus, we use two PID controllers to track roll and pitch angle respectively. 

\subsection{Results}
\begin{figure}[htbp]
\centering
\includegraphics[scale=0.35]{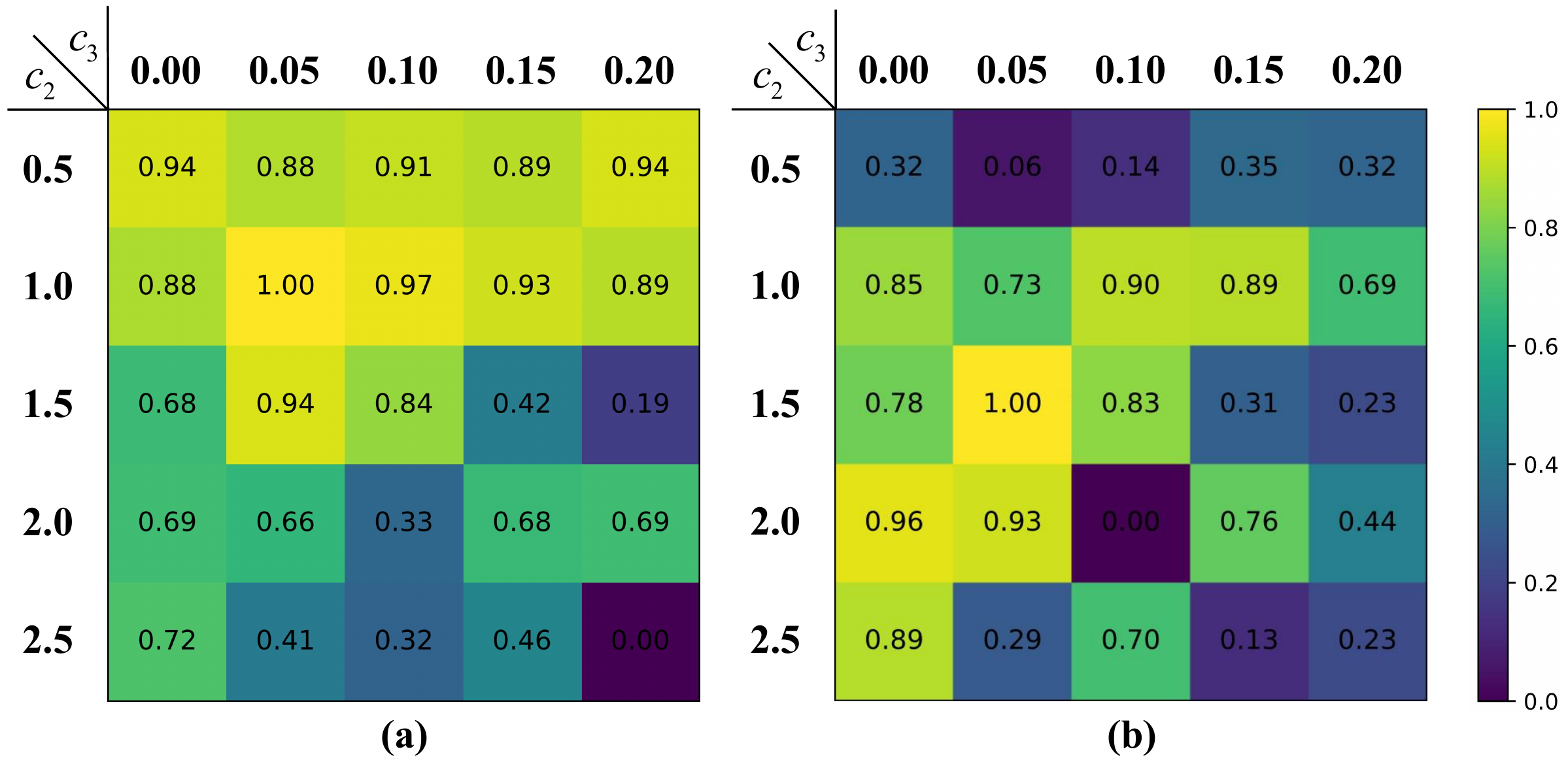}
\caption{Performance matrix of flight controller. (a) Alive bonus. The controller with higher score has better ability to keep stability. (b) Tracking performance. The controller with higher score can track target signals better. }
\label{fig:micro-matrix}
\end{figure}

As shown in Fig. \ref{fig:micro-matrix}, we train the flight controller with all pairs of coefficients, and test each controller with 64 random target signals. The evaluation metrics of Fig. \ref{fig:micro-matrix} (a) and (b) are the alive bonus and tracking error, respectively. The scores are normalized by their maximum and minimum values. A higher score means that the controller can better keep aircraft stability or track the target signals. Fig. \ref{fig:micro-matrix} shows that higher $c_2$ may lead to the instability of aircraft but higher tracking performance of flight controller. This is because that the flight controller focuses more on the tracking error and neglects to keep stability. Besides, small $c_3$ (0.05 and 0.10) can improve the performance of tracking signals and keeping stability. However, if $c_3$ continues to increase, the performance may decrease due to the much small deflection angle commands. 

\begin{figure}[ht]
\centering
\includegraphics[scale=0.45]{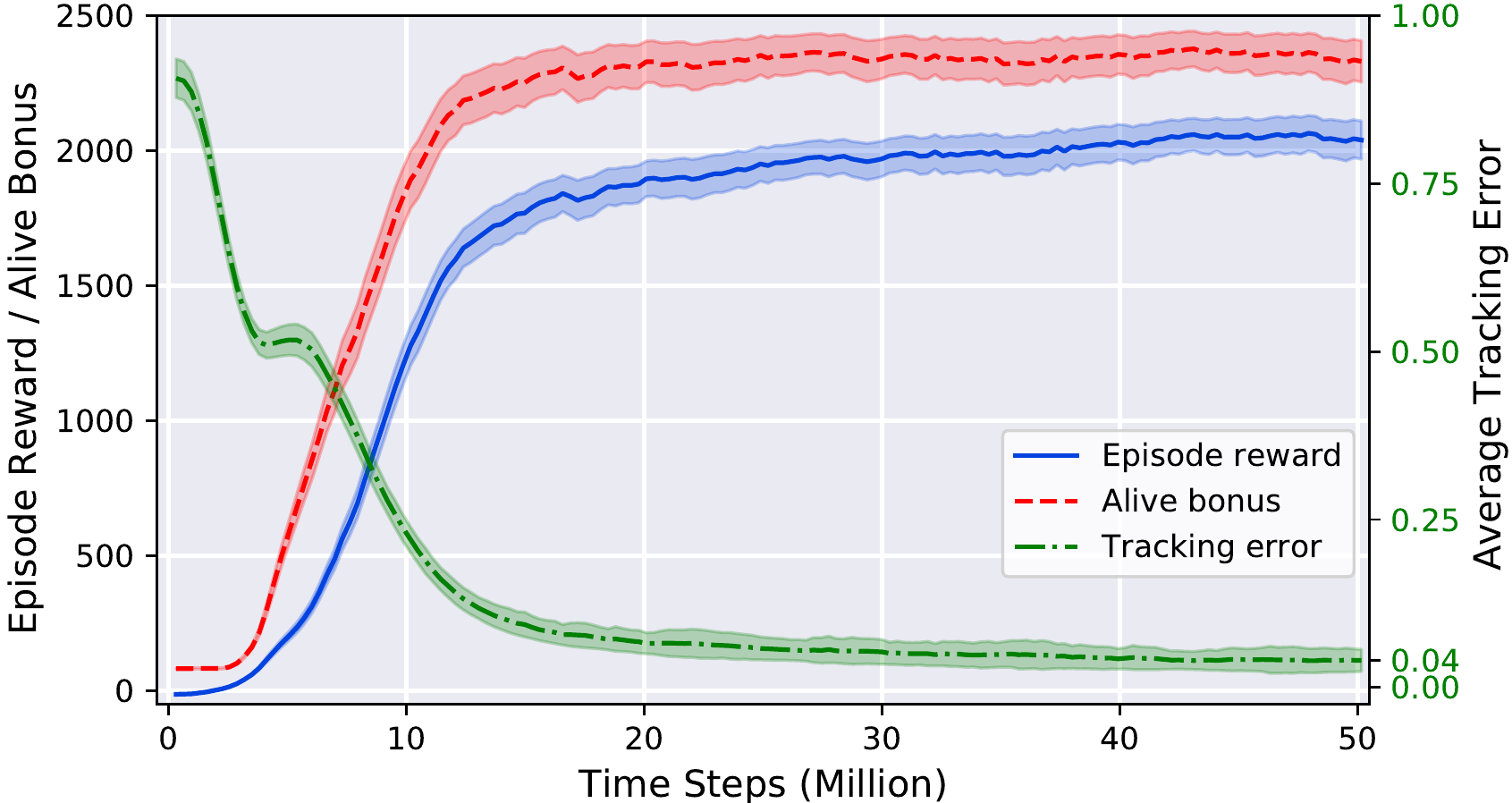}
\caption{Learning curves of the episode reward (blue), alive bonus (red), and tracking error (green) of the final flight controller.  }
\label{fig:micro-curves}
\end{figure}

\begin{figure*}[htbp]
\centering
\subfloat[]{
\begin{minipage}{5.75cm}
\centering
\includegraphics[scale=0.425]{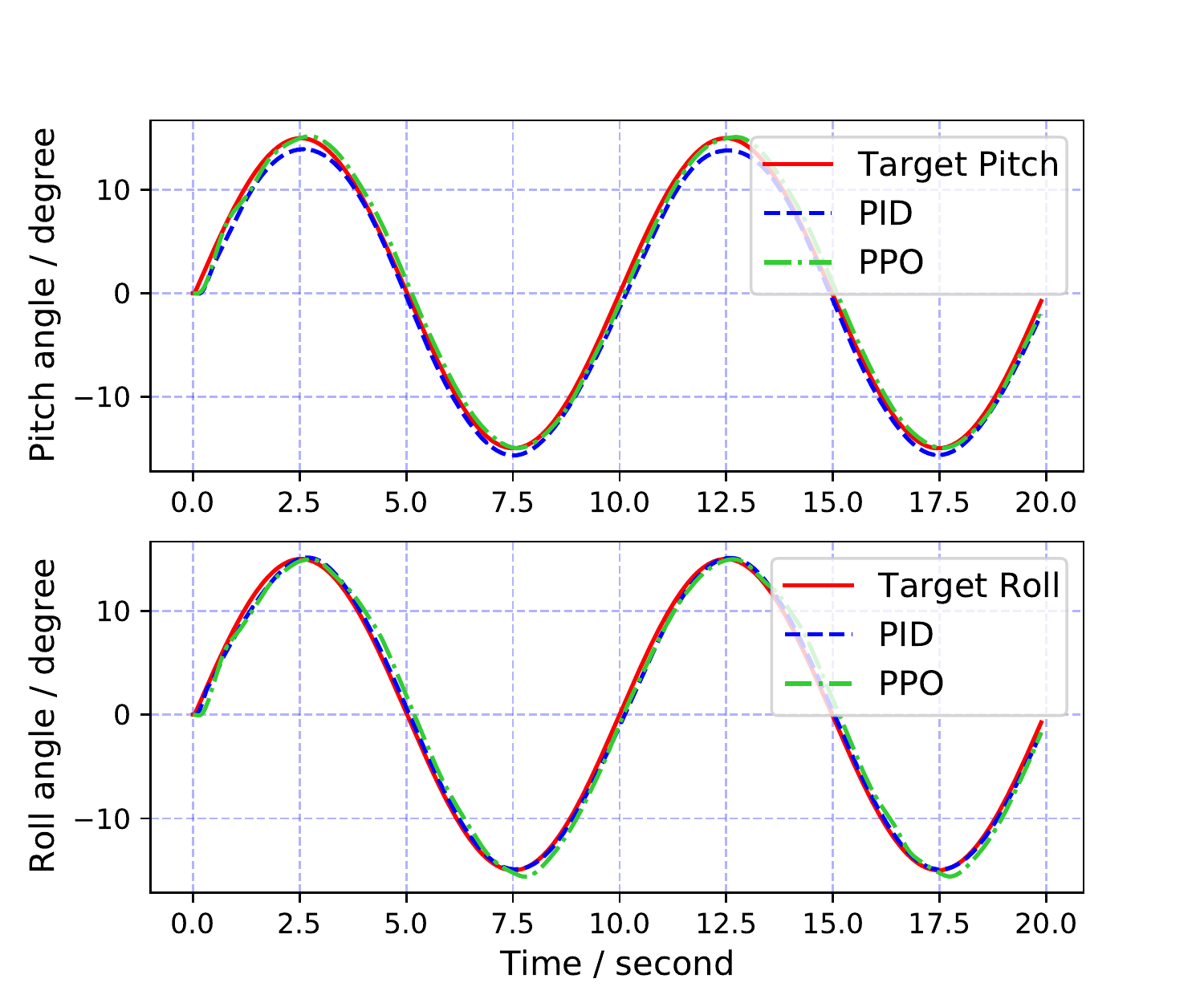}
\end{minipage}
}
\subfloat[]{
\begin{minipage}{5.75cm}
\centering
\includegraphics[scale=0.425]{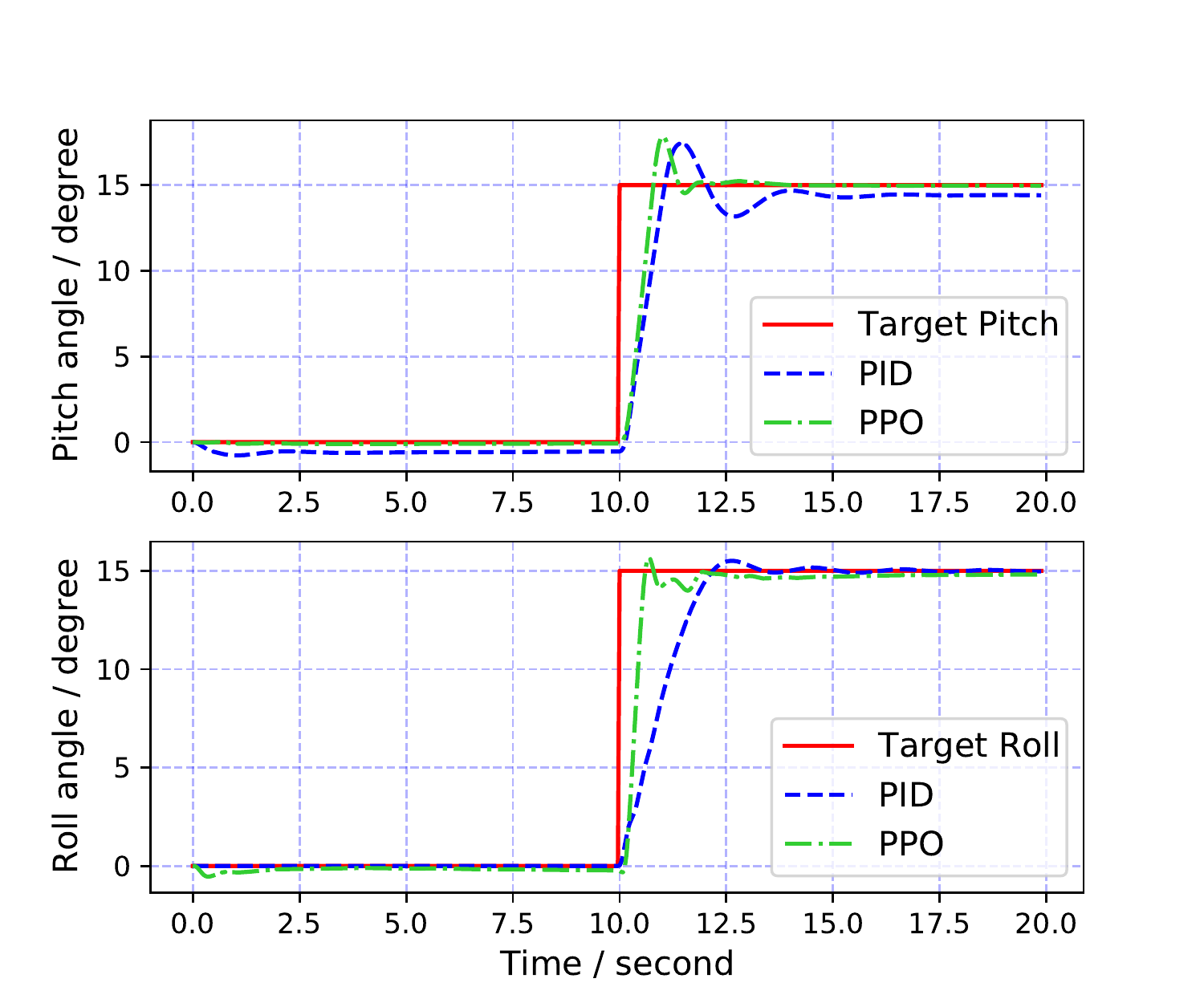}
\end{minipage}
}
\subfloat[]{
\begin{minipage}{5.75cm}
\centering
\includegraphics[scale=0.425]{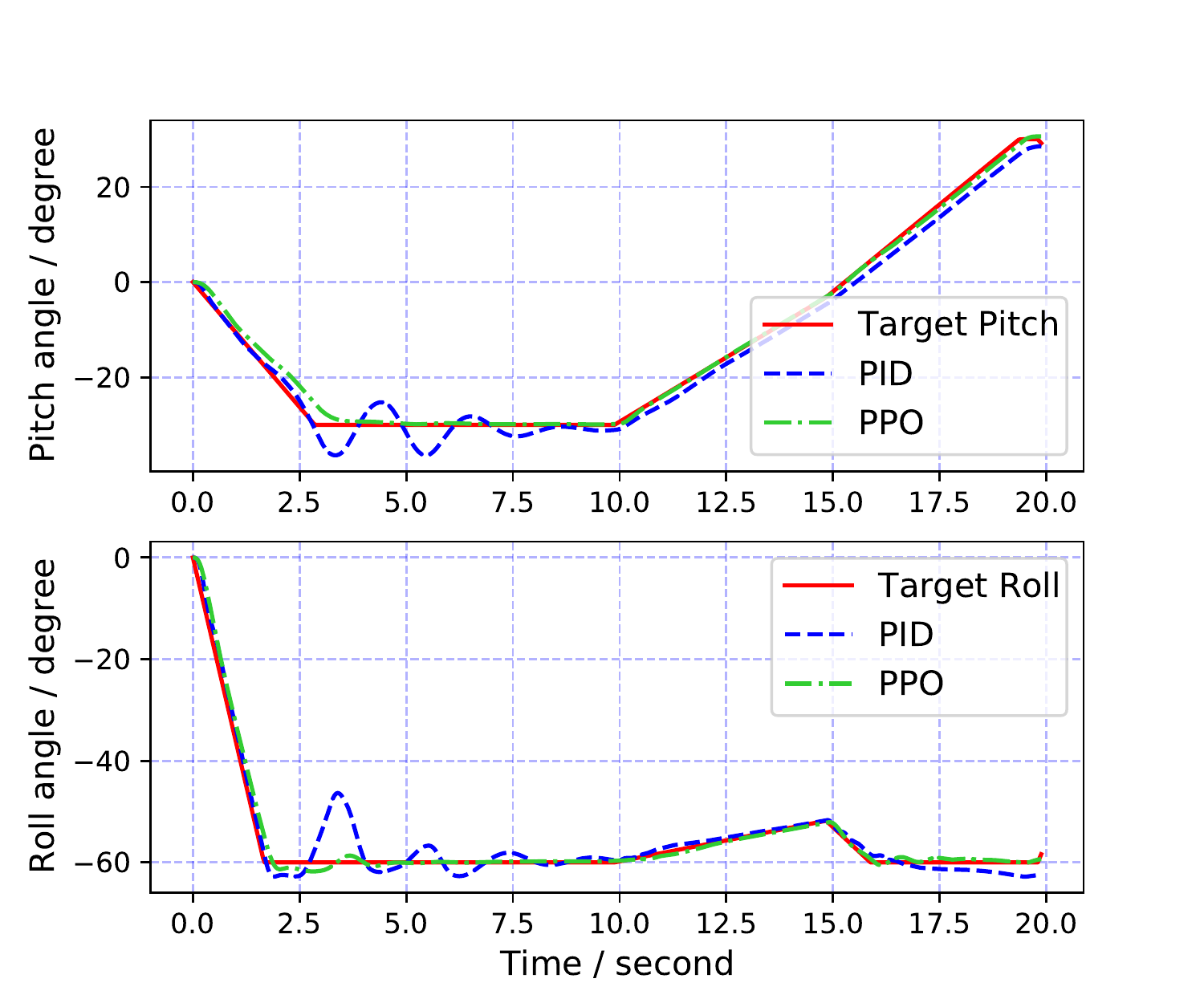}
\end{minipage}
}
\caption{Performance of trained PPO-based flight controller and PID-based flight controller. The controllers need to track the given target signals (roll and pitch angle). (a) sine-cosine signal. (b) step signal. (c) random signal. }
\label{fig:micro-results}
\end{figure*}

We choose the flight controller, whose training coefficients are $(c_2, c_3) = (1.5, 0.05)$, as the final flight controller. The learning curve is shown in Fig. \ref{fig:micro-curves}. The curves represent the mean values and the shaded parts represent the confidence intervals. The blue curve indicates that the episode reward rises rapidly at the beginning of the training, and then converges. The red curve and green curve represent the alive bonus and average tracking error of each episode, respectively. The two curves show that the stability and tracking performance increase with training timesteps. For the alive bonus curve, since we set the maximum timesteps of an episode as 2500 (100 seconds), the result indicates that the converged flight controller can stable the aircraft well. In the tracking error curve, the tracking error is the sum of roll and pitch error. The converged value (0.04) means that the flight controller can keep an average tracking error of 1.1 degrees. 

In order to show the tracking performance of the flight controller, we use the final flight controller to show the tracking trajectories in Fig. \ref{fig:micro-results}. We test it under the signals described in the last section. Besides, in order to get better performance of the PID controller, we fine tune their parameters. The fine-tuned parameters are shown in Table \ref{table:pid}. 

\begin{table}[htbp]
\centering
\caption{Coefficients of PID controllers}
\renewcommand{\arraystretch}{1.2}
\begin{tabular}{ccc}
\cline{1-3}
\diagbox[dir=SE]{Coefficient}{Controller}	& Roll Controller	& Pitch Controller 	\\ \cline{1-3}
$K_p$			& 0.2        	& 0.65			 \\
$K_i$			& 0.015        	& 0.08			 \\
$K_d$			& 12.0        	& 25.0		     \\
\cline{1-3}
\end{tabular}
\label{table:pid}
\end{table}

As shown in Fig. \ref{fig:micro-results}, both the PID controller and PPO-based controller can manipulate the aircraft to track the target signals. Compared with PID controller, PPO-based controller has smaller overshoot, faster response, and smaller steady-state error. The PID controller has obvious oscillation, while the PPO-based controller can still track the target signal quickly and smoothly, especially the step signal and random signal.

\section{Experiments on Air Combat Strategy}

\subsection{Experimental Setup}
In the fictitious self-play training, we train each generation strategy to combat against the former generations. We end one generation until the average reward does not increase significantly for more than 50 iterations, and the deviation is less than 10. Then, we assume that the obtained strategy is the approximate best response of the former generations. When visualizing the combating trajectories of combat strategies, We consider the following three scenarios with balancing advantages:
\begin{enumerate}[leftmargin = 1.5em]
\item[a)] \textbf{Face to Face}: The two aircraft face to face at the beginning. Since the two sides are rapidly approaching each other, this scenario can test the ability of air combat strategy in the face of danger. 

\item[b)] \textbf{Back to back}: The two aircraft back to back at the beginning. Since the two sides have enough time to make decisions, this scenario can better test the whole combating performance. 

\item[c)] \textbf{Parallel reverse}: At the beginning, the two aircraft have the same $pe$, but one flies east and the other west. This scenario is quite common in WVR air combat before the two sides are engaged in dogfight. It can test the ability of the combat strategy to accumulate advantages in the dogfight. 

\end{enumerate}

\subsection{Main Results}
\begin{figure}[b]
\centering
\includegraphics[scale=0.525]{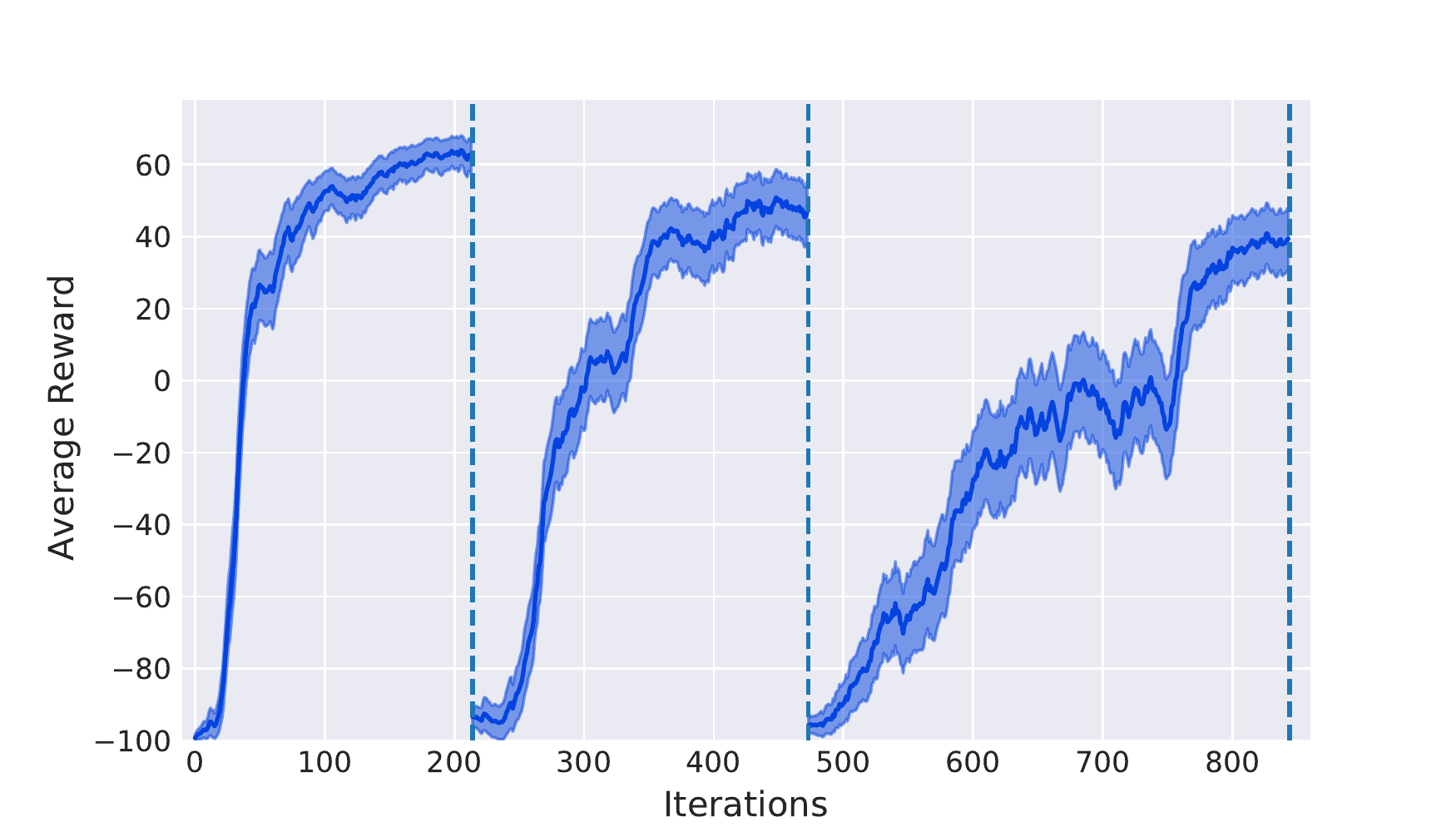}
\caption{Learning curves of the combat strategy via self-play. }
\label{fig:macro-curves}
\end{figure}
Based on the flight controller, we train the combat strategy from scratch via fictitious self-play. Fig. \ref{fig:macro-curves} shows the learning curve of the combat strategy. The vertical dashed lines break up the training curves of different generations, and we have a total of three generations of combat strategies. Within a generation, the average reward increases until convergence. Since the macro reward only indicates the outcome of the combat strategy, higher reward means that it can defeat the strategies in the historical strategy pool with better performance in various scenarios. Furthermore, the number of RL training iterations to converge increases with the number of generations. This is because the strategies in the pool become stronger, making it difficult to obtain an absolute advantage.

\begin{figure*}[htbp]
\centering
\subfloat{
\begin{minipage}{12cm}
\centering
\includegraphics[scale=0.35]{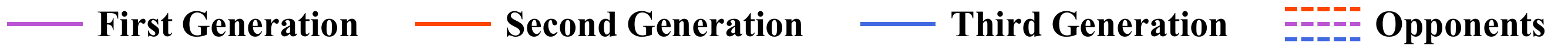}
\end{minipage}
}
\setcounter{subfigure}{0}
\subfloat[]{
\begin{minipage}{5.5cm}
\centering
\includegraphics[scale=0.07]{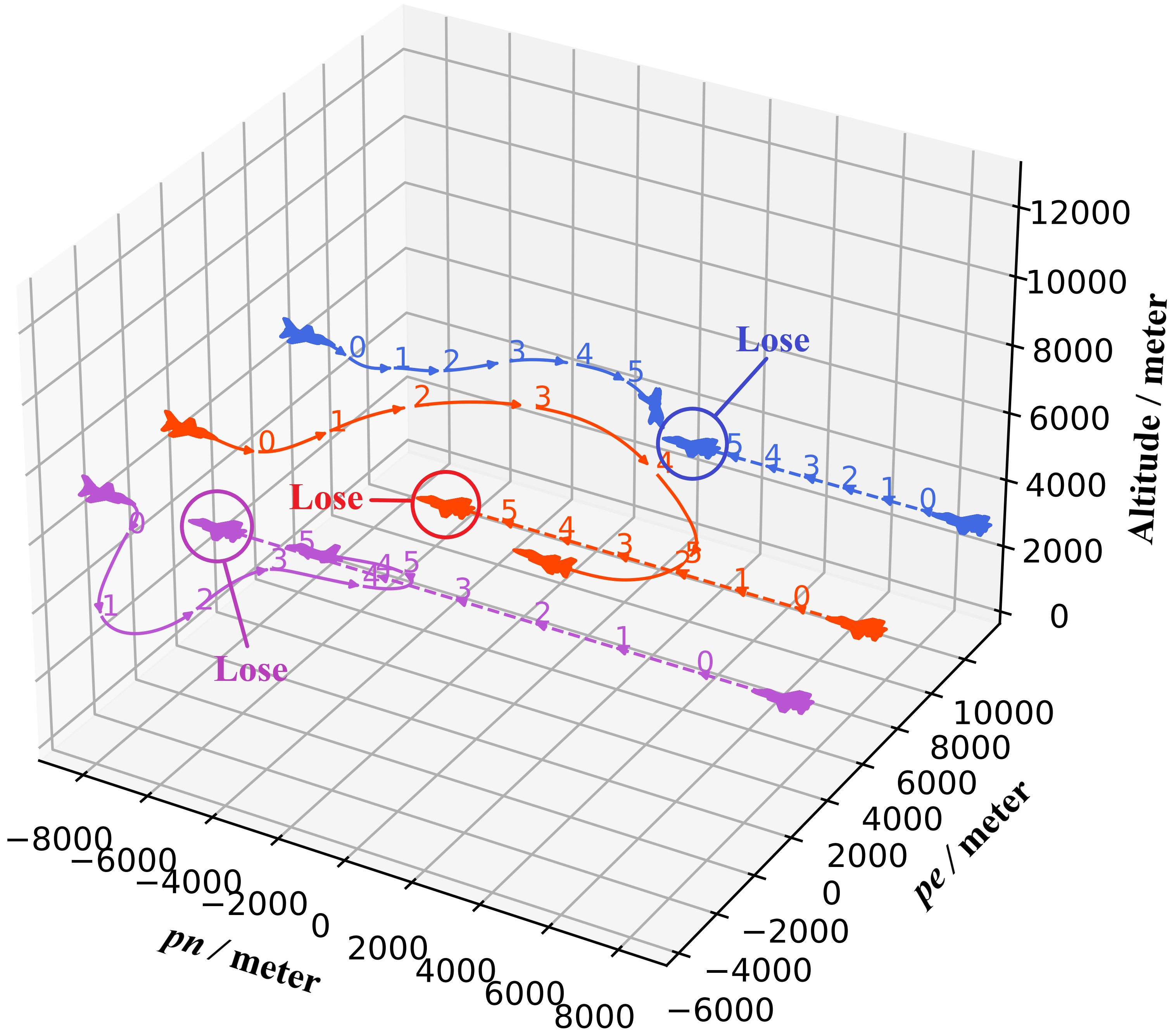}
\end{minipage}
}
\subfloat[]{
\begin{minipage}{5.5cm}
\centering
\includegraphics[scale=0.07]{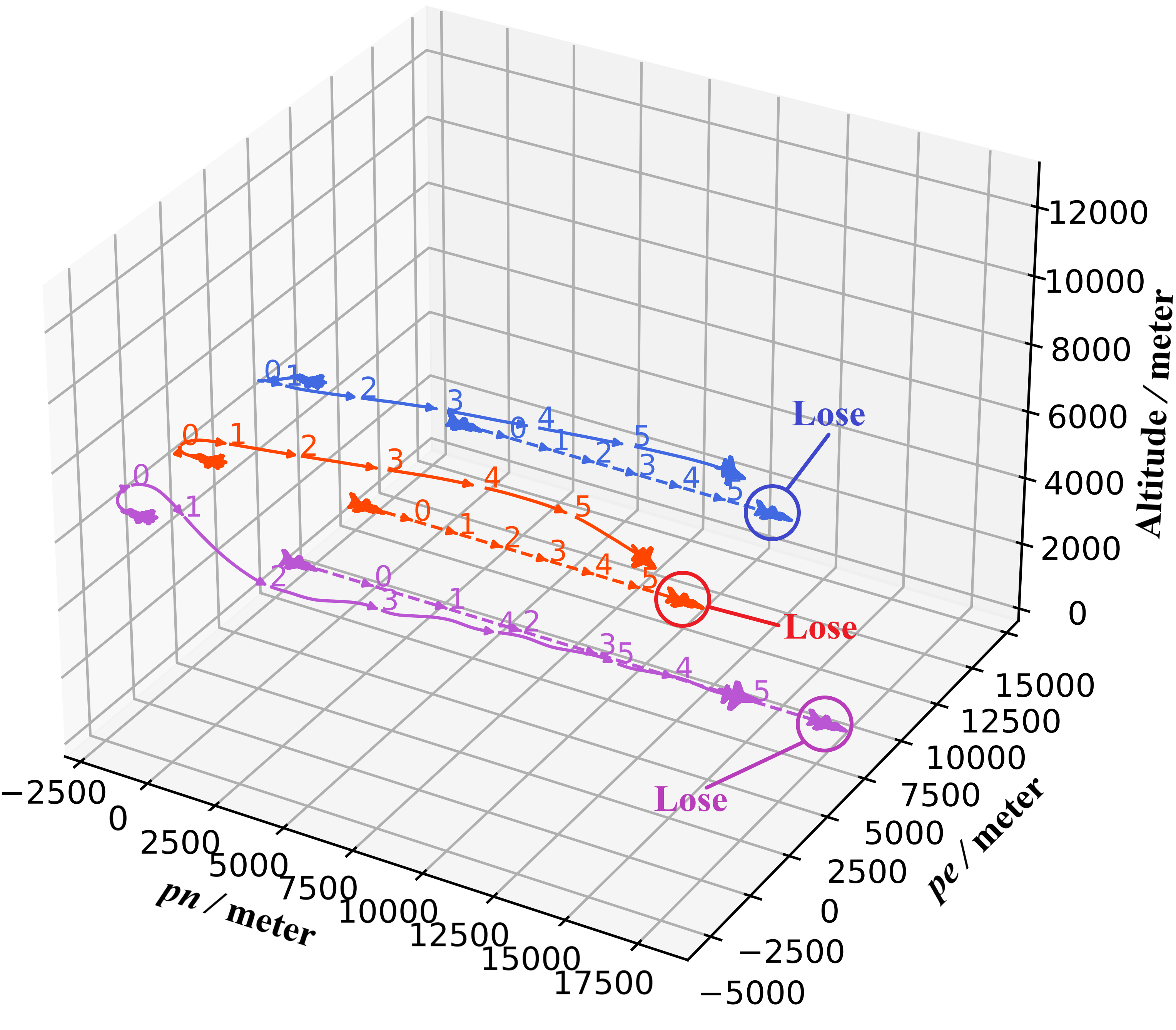}
\end{minipage}
}
\subfloat[]{
\begin{minipage}{5.5cm}
\centering
\includegraphics[scale=0.07]{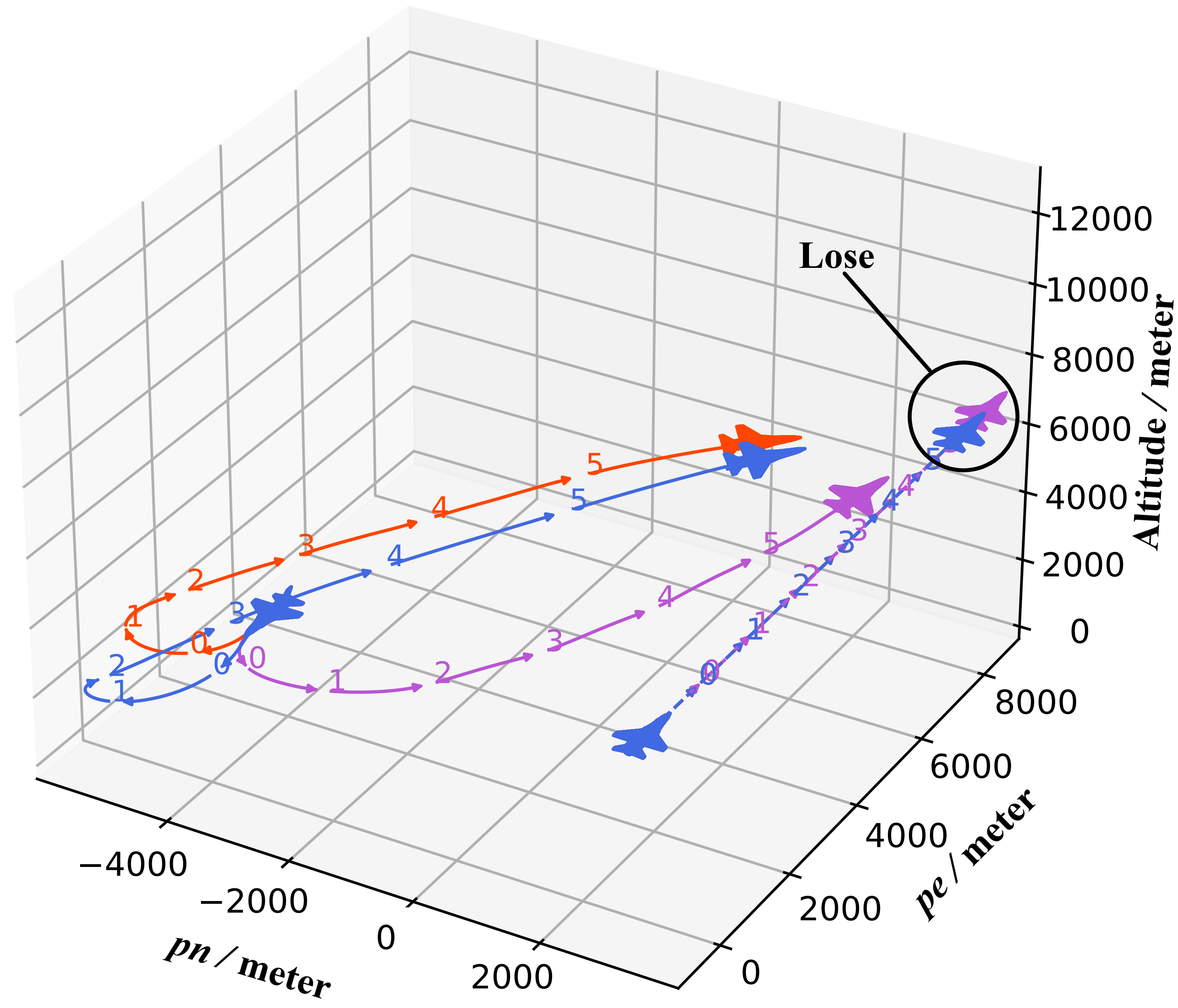}
\end{minipage}
}
\\
\subfloat[]{
\begin{minipage}{5.5cm}
\centering
\includegraphics[scale=0.07]{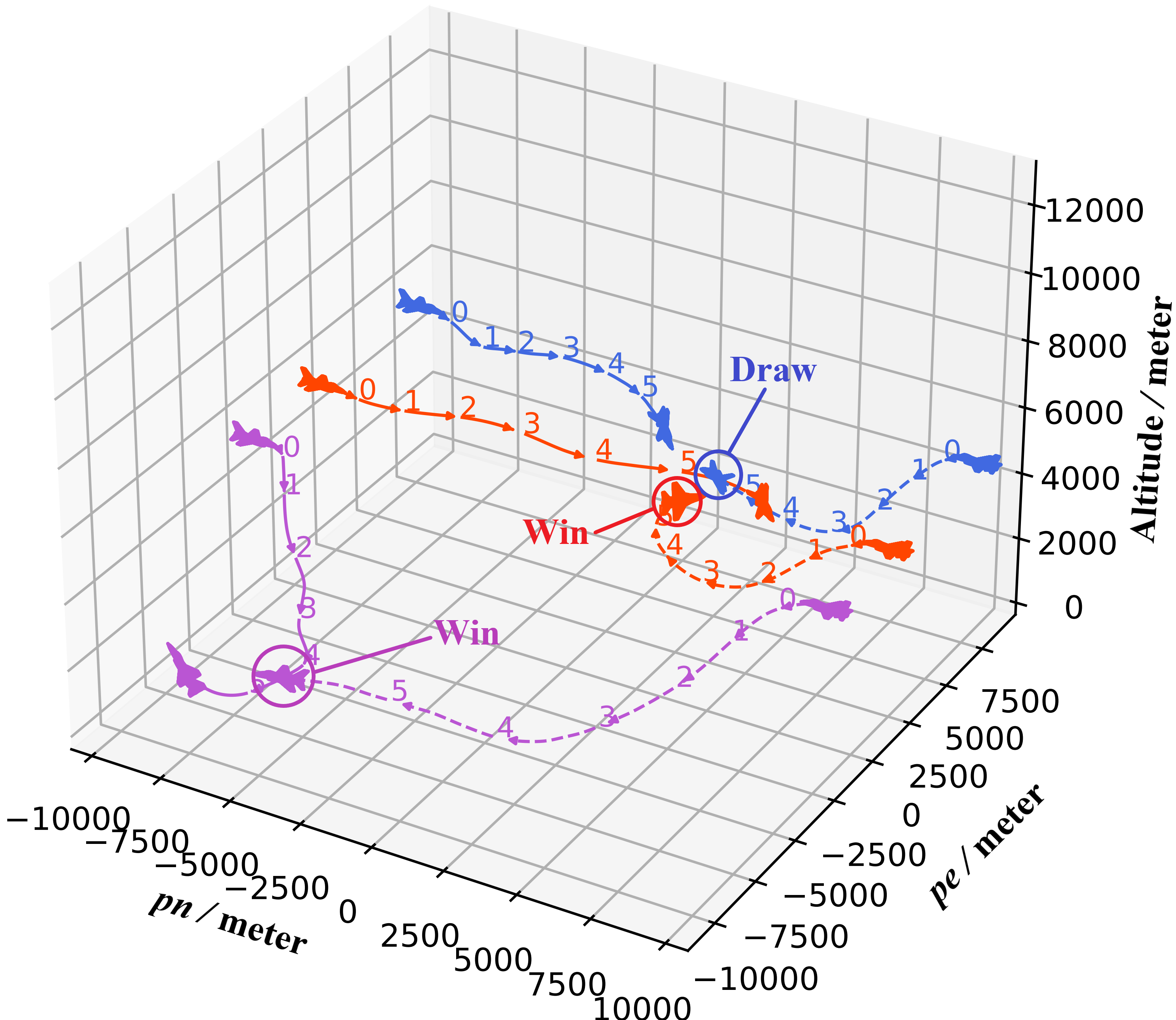}
\end{minipage}
}
\subfloat[]{
\begin{minipage}{5.5cm}
\centering
\includegraphics[scale=0.07]{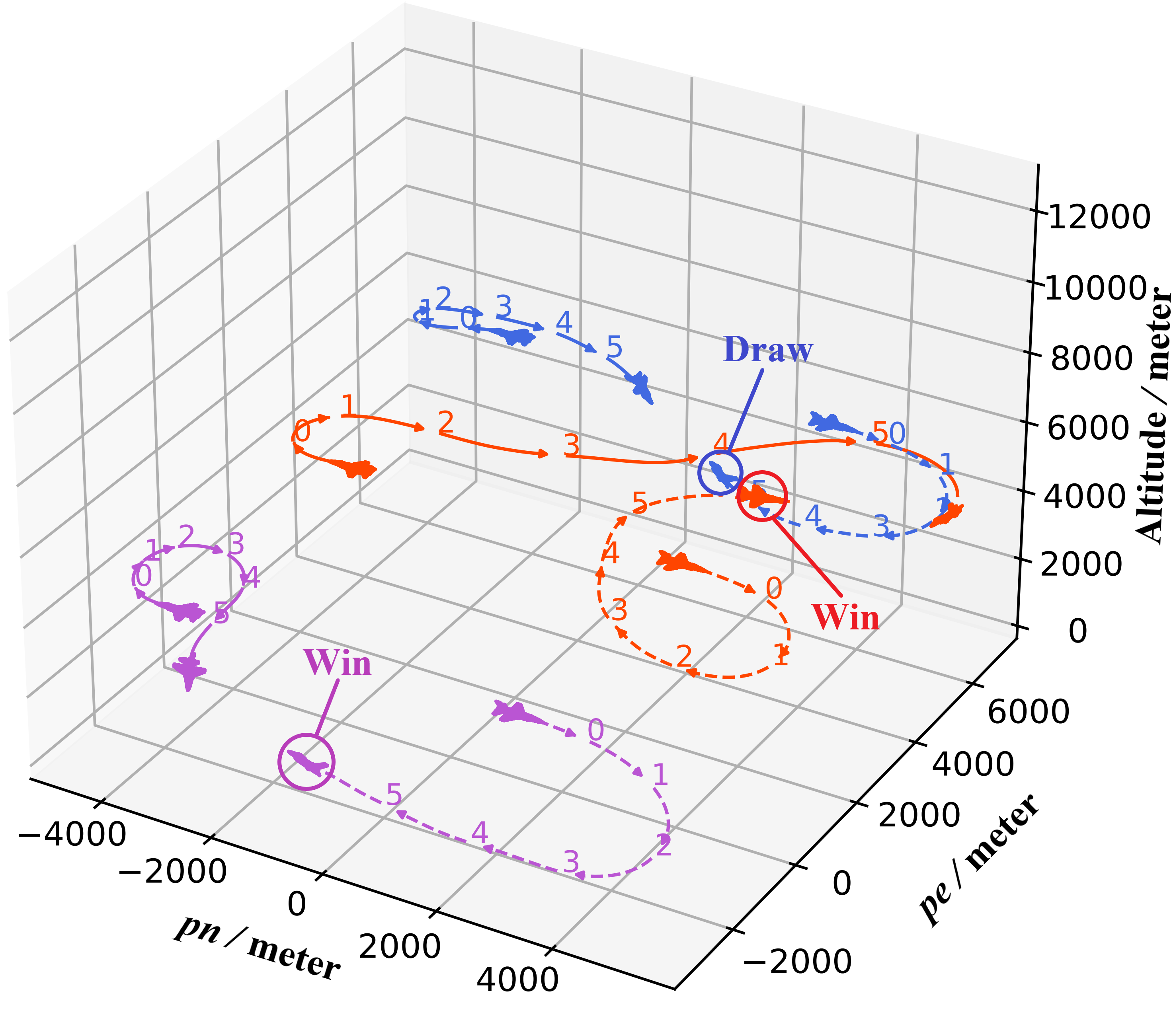}
\end{minipage}
}
\subfloat[]{
\begin{minipage}{5.5cm}
\centering
\includegraphics[scale=0.07]{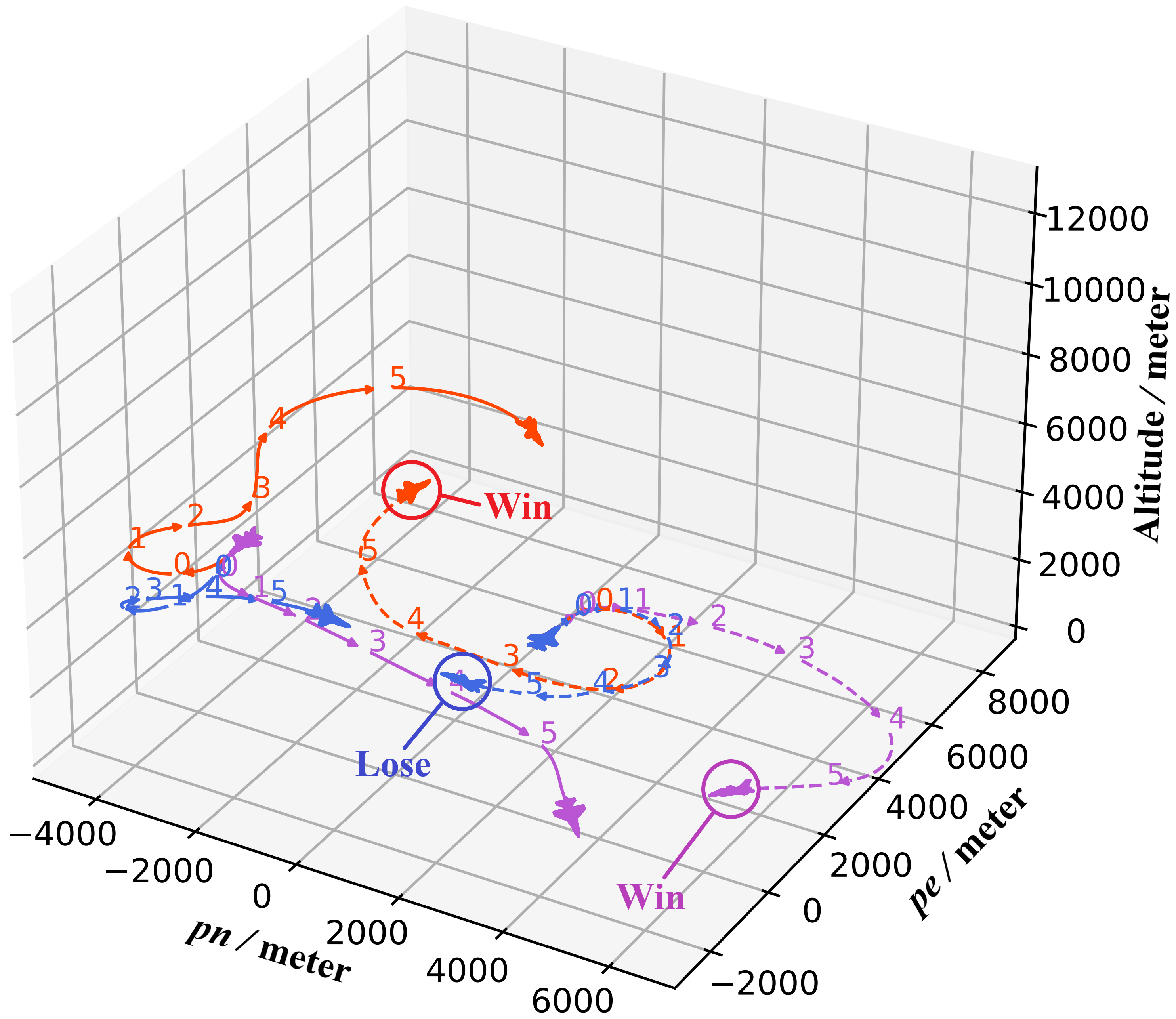}
\end{minipage}
}
\caption{Combating trajectories performed by the trained combat strategies of all generations. (a)(b)(c): Trajectories (solid) are performed by the combat trajectories at different generations of fictitious self-play against the level straight flight strategy (dashed). (d)(e)(f): Trajectories (solid) performed by all generation combat strategy against the third generation combat strategy (dashed). We mark the combat results of the opponents (dashed trajectories) in all figures. }
\label{fig:macro-trajectories}
\end{figure*}

\begin{table}[htbp]
\centering
\caption{Winning Rates of 1P fighting against 2P}
\renewcommand{\arraystretch}{1.2}
\begin{tabular}{c|ccc|c}
\cline{1-5}
\diagbox[dir=SE]{1 P}{2 P}	& level straight	& 1st-Gen 	& 2nd-Gen   & 3nd-Gen (RL) 	\\ \cline{1-5}
1st-Gen			& 0.859        	& -				& -			& - \\
2nd-Gen			& 0.953        	& 0.875			& -			& - \\
3rd-Gen			& 0.938        	& 0.781			& 0.844		& - \\
\cline{1-5}
3rd-Gen (PID)   & 0.578         & 0.500         & 0.313     & 0.343 \\
\cline{1-5}
\end{tabular}
\begin{tablenotes}
\item[1] * The first part is the main experiment results, and the second part is the ablation experiment results. 
\end{tablenotes}
\label{table:win-rates}
\end{table}

In order to show the improvement between each generations, we test the winning rates between each pair of strategies by performing 64 combats. As shown in Table \ref{table:win-rates}, higher-generation strategies can get higher winning rates against the lower-generation strategies. It demonstrates that adding stronger strategies generation by generation in the fictitious self-play process can further improve the performance of learned strategies. Besides, since the higher generation strategy is trained by fighting against the uniformly sampled strategy pool, its winning rate against a specific strategy may decline in order to ensure a high overall winning rate against all strategies. 

As shown in Fig. \ref{fig:macro-trajectories}, we plot the combating trajectories in the scenarios described in the last section to better demonstrate the effect of the self-play mechanism. We plot the aircraft at the start and end positions to show the situation. Besides, in order to show more details of the combat, we divide the trajectories into several equal parts and mark the corresponding numbers. The same numbers of the two trajectories correspond to the same timestep. In Fig. \ref{fig:macro-trajectories}, trajectories with the same color represent the same combat, and dashed trajectories represent the opponents. The results of each combat are marked in the figure. 

\begin{enumerate}[leftmargin = 1.5em]
\item[1)] \textbf{Against Level Straight Flight}: Fig. \ref{fig:macro-trajectories} (a)(b)(c) show combating trajectories performed by all generation strategies against the level straight flight strategy. As shown in Fig. \ref{fig:macro-trajectories} (a), the first and second generations have to conduct long-distance maneuvers to avoid being attacked, and attack the opponent from the back. However, the third generation strategy can accurately bypass the opponent's attack area and consume the least time to defeat it. In the second scenario shown in Fig. \ref{fig:macro-trajectories} (b), all strategies can quickly execute 180-degree turns, and strategies with higher generation can turn faster. Fig. \ref{fig:macro-trajectories} (c) also demonstrates this. 

\item[2)] \textbf{Against Third Generation}: Fig. \ref{fig:macro-trajectories} (d)(e)(f) show combating trajectories performed by all generation strategies against the third generation strategy. In all scenarios, the third generation strategy shows a strong aggressiveness, which tends to quickly aim its nose direction at the opponent. It learns a better maneuver to change its flight direction faster than its opponents to gain more advantages. Therefore, all trajectories end with the third generation opponent winning the combat. Besides, compared with the first and second generation, the trajectories of the third generation is smoother, which means that it is more stable. 

\end{enumerate}

\subsection{Strategy Analysis}
\begin{figure*}[htbp]
\centering
\includegraphics[scale=0.525]{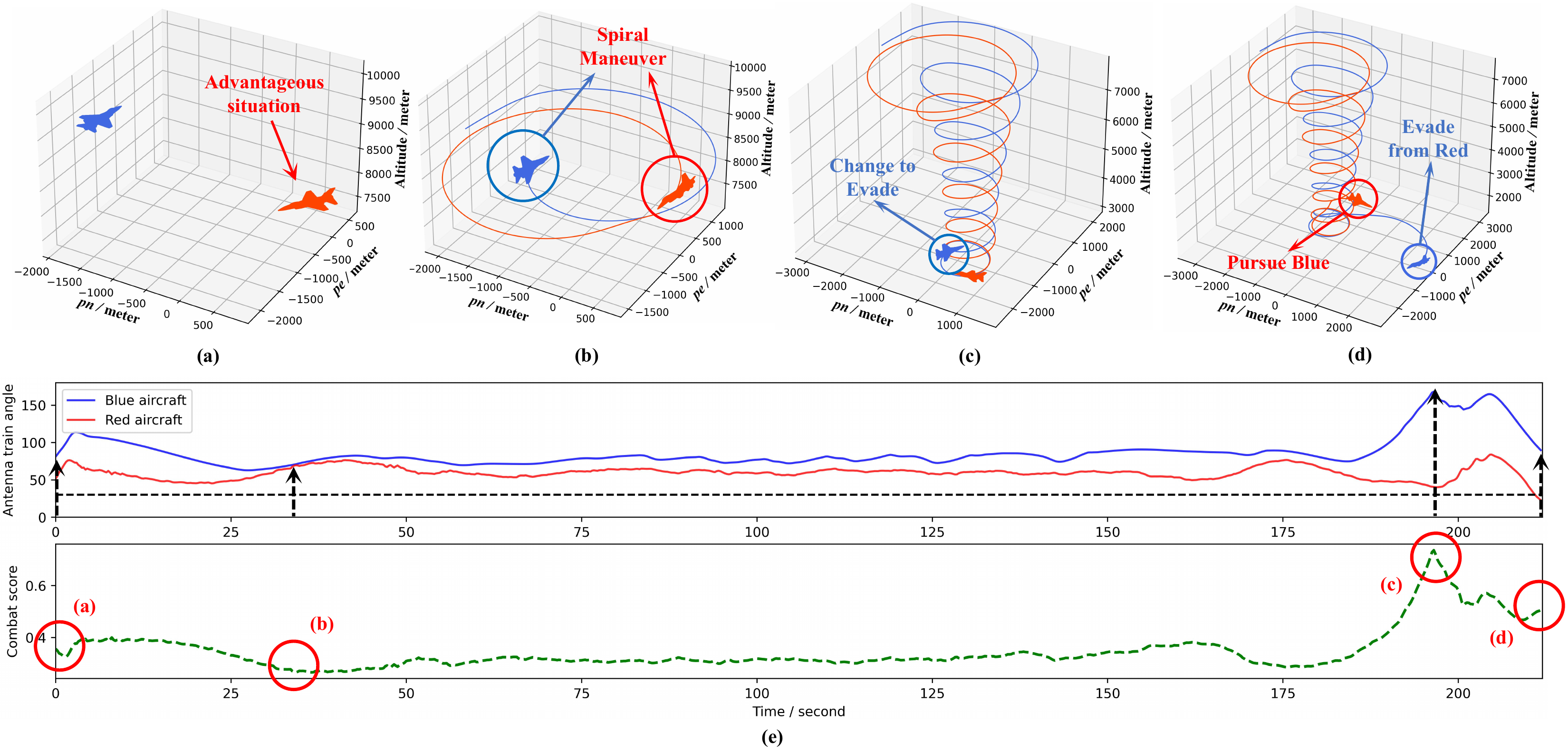}
\caption{The detail of a trajectory performed by the third generation strategy. (a) Initial situation that the red aircraft has advantage. (b) both sides are performing a spiral maneuver. (c) the blue aircraft is in danger, so it changes tactics and chooses to evade from the red aircraft. (d) the blue aircraft is eventually shot down. (e) the combat score and the ATAs during this combat. }
\label{fig:detail-maneuver}
\end{figure*}
Next, we show a typical maneuver in Fig. \ref{fig:detail-maneuver}, where both the red and blue aircraft are controlled by the third generation strategy. In order to better analyze the learned maneuvers by the third generation strategy, we use the score function proposed in \cite{shin2018autonomous} to evaluate the situation at all timesteps. The score function from the perspective of red aircraft is described by the ATAs of red and blue aircraft as follows:
\begin{figure}[ht]
\centering
\includegraphics[scale=0.075]{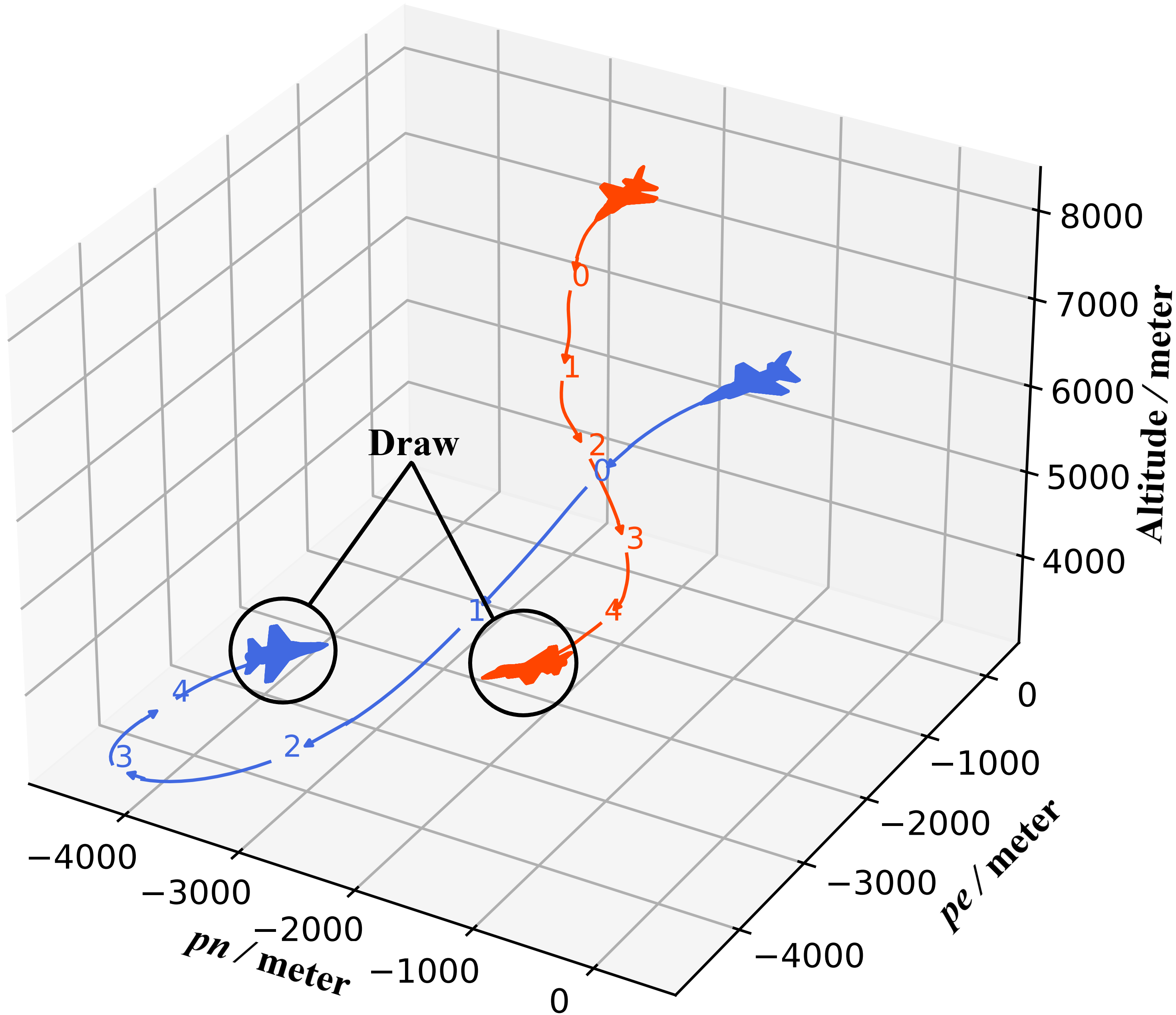}
\caption{Trajectories performed by the third generation combat strategy. The blue aircraft performs a sharp turn to avoid being shot down. }
\label{fig:other-trajectories}
\end{figure}
\begin{equation}
Score = \frac{\mu(\pi-|\omega_a^r|)^2 + (1-\mu)|\omega_a^b|^2}{\pi^2} \in [0, 1]
\end{equation}
where $\mu$ is a hyper-parameter, which is set to 0.5 according to the suggestion given in \cite{shin2018autonomous}. $\omega_a^r$ and $\omega_a^b$ are the ATAs of red and blue aircraft, respectively. This score varies from 0 to 1. It takes 1 when the red aircraft is directly behind the blue aircraft, and 0 on the contrary. Besides, it is easy to calculate that the score is 0.25 when the two aircraft are in a completely neutral situation, where all ATAs are $\pi / 2$. Fig. \ref{fig:detail-maneuver} (e) shows the ATAs and combat score of this trajectory from the perspective of red aircraft. Smaller $\omega_a^r$ and larger $\omega_a^b$ result in larger score. We choose four typical frames to analyze the whole trajectory as show in Fig. \ref{fig:detail-maneuver} (a)-(d). As shown in Fig. \ref{fig:detail-maneuver} (a), the red aircraft has an advantageous situation at the beginning, where its ATA is smaller than that of the blue aircraft. Fig. \ref{fig:detail-maneuver} (b) shows a frame with the smallest combat score for the red aircraft. Both aircraft have similar ATAs, so that they try to accumulate advantages through spiral maneuver. Then, they keep narrowing their turning radius, trying to get behind its enemy. At the end of spiral maneuver, the red aircraft accumulates enough advantages and almost targets the blue aircraft in its attack area as shown in Fig. \ref{fig:detail-maneuver} (c). Due to the disadvantage of the blue aircraft, it decides to change its tactics to evade from the red aircraft instead of chasing it. However, it is eventually shot down by the red aircraft as shown in Fig. \ref{fig:detail-maneuver} (d). 

Besides, there are also other typical trajectories as shown in Fig. \ref{fig:other-trajectories}, where both aircraft are also controlled by the third generation strategy. It is a typical pursuit-evasion scenario, where the red aircraft is behind its enemy at the initial timestep. The red aircraft keeps pursuing behind its enemy, while the blue aircraft performs a sharp turn after getting an appropriate distance from the red aircraft. The result of this combat is that both sides reach a draw with mutual destruction. As we described above, the third generation strategy has strong aggressiveness. The aircraft does its best to target its enemy in the attack area, which led to this combat result. 

\subsection{Decision Experiment}
\begin{figure}[ht]
\centering
\includegraphics[scale=0.08]{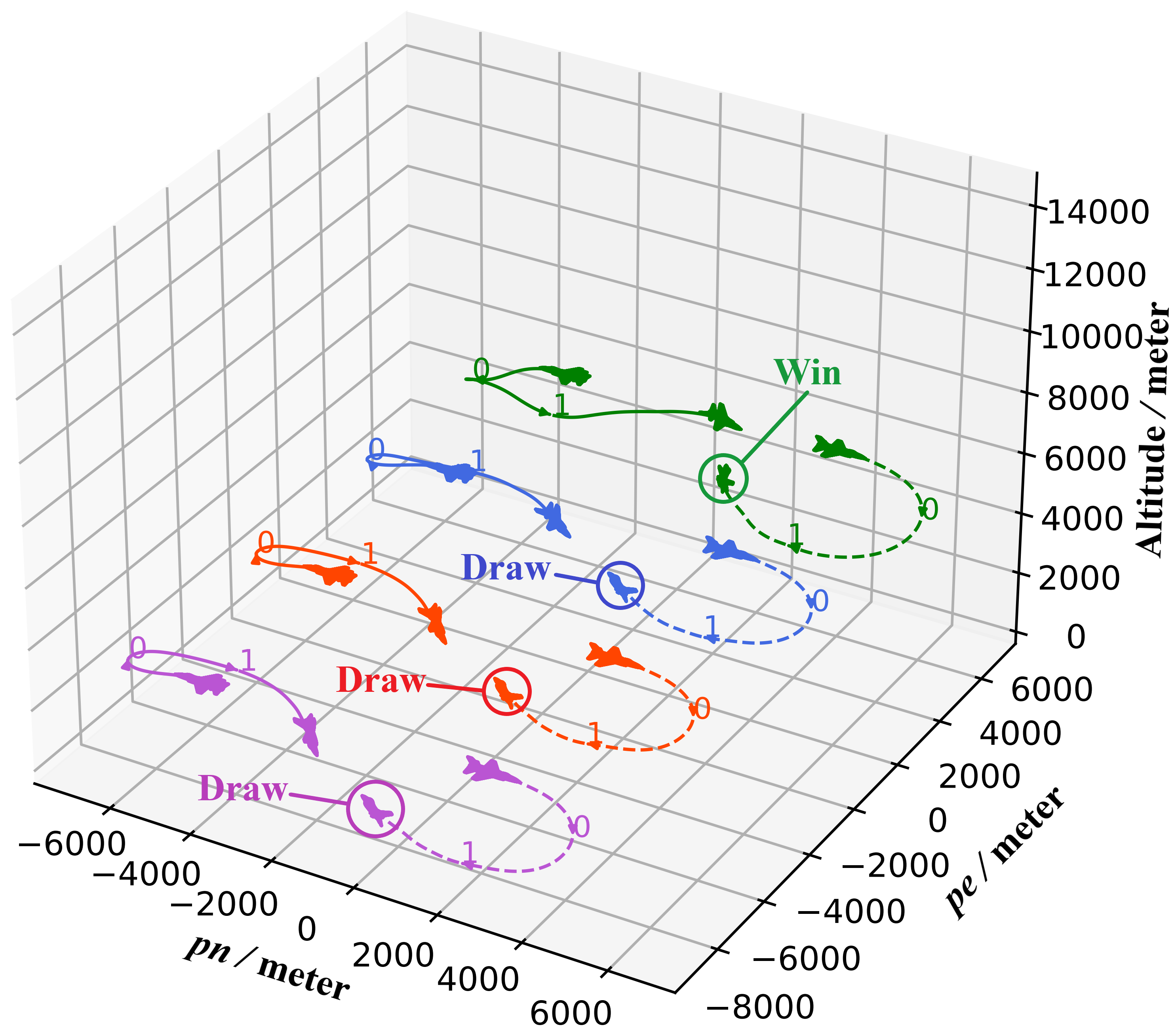}
\caption{Trajectories performed by the third generation combat strategy with different decision intervals. The decision intervals of dashed trajectories are always 0.25 seconds. The decision intervals of solid are 0.1 seconds (purple), 0.25 seconds (red), 1.0 seconds (blue), and 4.0 seconds (green). }
\label{fig:decision-trajectories}
\end{figure}

We also investigate the effect of the decision frequency on the performance of combat strategies. As shown in Fig. \ref{fig:decision-trajectories}, both sides use the third generation combat strategy. The dashed trajectories are performed by the strategy whose decision interval is 0.25 seconds. The solid trajectories are performed by the strategies whose decision intervals are 0.1 seconds (purple), 0.25 seconds (red), 1.0 seconds (blue), and 4.0 seconds (green). When the decision interval is 0.1 seconds, 0.25 seconds and 1.0 seconds, both sides reach a draw with mutual destruction. When the decision interval is further increased to 4.0 seconds, the aircraft is shot down by its enemy. This is because that the strategy is unable to perform timely response at some critical timesteps, making it continuously lose its advantage. The results show that the much small decision frequency (high decision interval) may lead to the decrease of strategy performance. 

\subsection{Ablation Experiment}
In the ablation experiments, we present two parts to illustrate the effect of our method. 
\begin{enumerate}[leftmargin = 1.5em]
\item[1)] \textbf{Replace the RL-based flight controller}: In the proposed framework, the combat strategy depends on the RL-based flight controller, which has better tracking performance than PID controller. In this experiment, we replace the RL-based flight controller with the PID controller to test their winning rates. The third generation combat strategy with RL-based controller fights against all strategies with PID controller as shown in the second part of Table \ref{table:win-rates}. We also get these winning rates by performing 64 random combats. The winning rate of the combat strategy with PID controller is much lower than the strategy with RL-based controller. Fig. \ref{fig:ablation-result} shows the trajectories of one back to back combat. The blue trajectory is performed by the combat strategy with RL-based controller, while the red trajectory is performed by the combat strategy with PID controller. The turning speed of the red aircraft is obviously lower than the blue aircraft, which leads to its failure. On the one hand, this result indicates that the RL-based flight controller can better execute the macro commands from the combat strategy, thus gaining more combat advantages. On the other hand, it also demonstrates that the proposed combat strategy can be compatible with other flight controllers and achieve intelligent control. 

\begin{figure}[ht]
\centering
\includegraphics[scale=0.08]{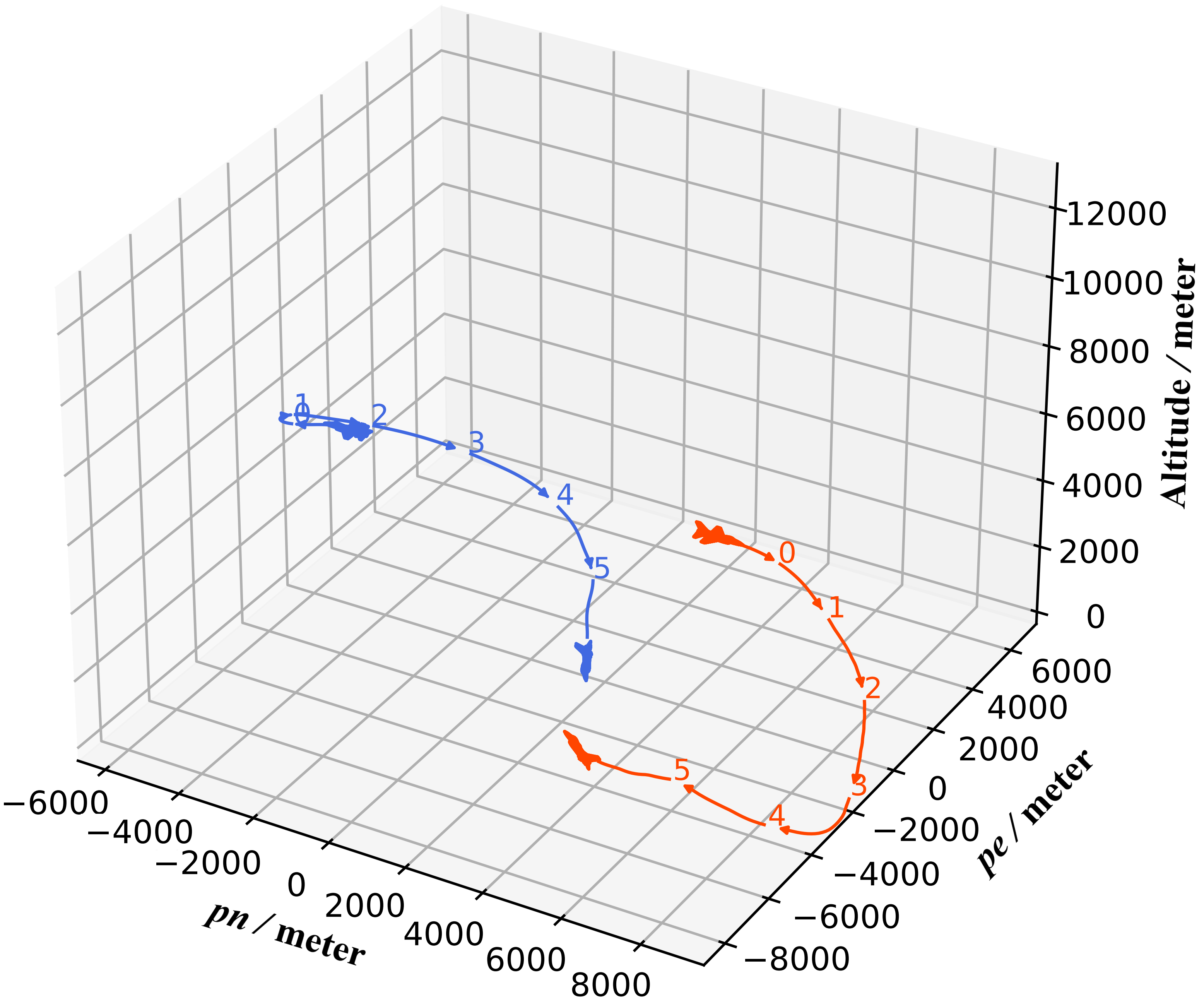}
\caption{Trajectories performed by the combat strategy with RL-based flight controller (blue) and PID-based flight controller (red). }
\label{fig:ablation-result}
\end{figure}

\item[2)] \textbf{Remove fictitious self-play mechanism}: The first generation strategy is actually the result of the purely PPO training experiment, which is an ablation experiment to remove the fictitious self-play training. The higher generation strategies can achieve more than 75\% winning rate against it. This result shows that the strategies trained by fictitious self-play mechanism can achieve better performance than the strategies trained by simple RL algorithms. 

\end{enumerate}

\section{Conclusion}
This article presents a novel training framework for WVR air-to-air combat under 6-DOF dynamics. We propose a hierarchical framework for aircraft control. The flight controller holds the inner loop and the combat strategy holds the outer loop. Both of them are trained by PPO algorithm. Furthermore, we employ the self-play mechanism to improve the strategy performance. Experiments results demonstrate the performance of the flight controller and combat strategy. The flight controller can control the aircraft to track several kinds of target signals quickly and accurately, and the combat strategy can learn powerful tactics to defeat enemies. We also demonstrate the effect of the evolution between different generations and analyze the learned maneuver behaviors of the third generation strategy. These results show that the proposed training framework can address the difficult 6-DOF air combat problem. Besides, the proposed framework can also help to solve other continuous action space zero-sum games. In the future, general-sum games, such as multi-aircraft air combat, can be considered to expand the application of this framework.


%

\bibliographystyle{IEEEtran}
\bibliography{IEEEabrv, ref}




\end{document}